\PassOptionsToPackage{table}{xcolor}
\documentclass[acmlarge]{acmart}
\usepackage{amsmath,amsfonts}
\usepackage{algorithmic}
\usepackage{array}
\usepackage{textcomp}
\usepackage{stfloats}
\usepackage{url}
\usepackage{verbatim}
\usepackage{graphicx}
\usepackage{adjustbox}
\usepackage{caption}
\usepackage{subcaption}
\usepackage{multirow}
\usepackage{fancyref}
\usepackage{color}
\usepackage{float} % to force float with [H]
\usepackage{algorithm2e}
\usepackage[capitalize]{./cleveref}
\usepackage{listings}

\usepackage{pifont}% http://ctan.org/pkg/pifont
\newcommand{\cmark}{\ding{51}}%
\newcommand{\xmark}{\ding{55}}%
\definecolor{turquoise}{cmyk}{0.65,0,0.1,0.3}
\definecolor{purple}{rgb}{0.65,0,0.65}
\definecolor{dark_green}{rgb}{0, 0.5, 0}
\definecolor{orange}{rgb}{0.8, 0.6, 0.2}
\definecolor{red}{rgb}{0.8, 0.2, 0.2}
\definecolor{darkred}{rgb}{0.6, 0.1, 0.05}
\definecolor{blueish}{rgb}{0.0, 0.3, .6}
\definecolor{light_gray}{rgb}{0.7, 0.7, .7}
\definecolor{pink}{rgb}{1, 0, 1}
\definecolor{greyblue}{rgb}{0.25, 0.25, 1}
\definecolor{LightCyan}{rgb}{0.88,0.95,1}

\definecolor{plotviolet}{RGB}{149, 108, 180}
\definecolor{plotred}{RGB}{214, 95, 95}
\definecolor{plotgreen}{RGB}{106, 204, 100}
\definecolor{plotdarkgreen}{HTML}{99C09E}

\newcommand{\hide}[1]{}

\def\BibTeX{{\rm B\kern-.05em{\sc i\kern-.025em b}\kern-.08em
    T\kern-.1667em\lower.7ex\hbox{E}\kern-.125emX}}
\usepackage{balance}

\setcopyright{acmcopyright}
\acmJournal{TOMM}

\begin{document}
\title{Composed Image Retrieval using Contrastive Learning and Task-oriented CLIP-based Features}
\author{Alberto Baldrati}
\email{alberto.baldrati@unifi.it}
\affiliation{%
  \institution{Università degli Studi di Firenze - MICC}
  \city{Firenze}
  \country{Italy}
}
\affiliation{%
  \institution{Università di Pisa}
  \city{Pisa}
  \country{Italy}
}
\author{Marco Bertini}
\email{marco.bertini@unifi.it}
\affiliation{%
  \institution{Università degli Studi di Firenze - MICC}
  \city{Firenze}
  \country{Italy}
}
\author{Tiberio Uricchio}
\email{tiberio.uricchio@unimc.it}
\affiliation{%
  \institution{Università degli Studi di Macerata}
  \city{Macerata}
  \country{Italy}
}
\author{Alberto Del Bimbo}
\email{alberto.delbimbo@unifi.it}
\affiliation{%
  \institution{Università degli Studi di Firenze - MICC}
  \city{Firenze}
  \country{Italy}
}

\begin{abstract}
Given a query composed of a reference image and a relative caption, the Composed Image Retrieval goal is to retrieve images visually similar to the reference one that integrates the modifications expressed by the caption. Given that recent research has demonstrated the efficacy of large-scale vision and language pre-trained (VLP) models in various tasks, we rely on features from the OpenAI CLIP model to tackle the considered task. We initially perform a task-oriented fine-tuning of both CLIP encoders using the element-wise sum of visual and textual features. Then, in the second stage, we train a Combiner network that learns to combine the image-text features integrating the bimodal information and providing combined features used to perform the retrieval. We use contrastive learning in both stages of training. 
Starting from the bare CLIP features as a baseline, experimental results show that the task-oriented fine-tuning and the carefully crafted Combiner network are highly effective and outperform more complex state-of-the-art approaches on FashionIQ and CIRR, two popular and challenging datasets for composed image retrieval. Code and pre-trained models are available at \href{https://github.com/ABaldrati/CLIP4Cir}{\url{https://github.com/ABaldrati/CLIP4Cir}}

\end{abstract}

	

\maketitle

% \begin{IEEEkeywords}

% \end{IEEEkeywords}

\section{Introduction}\label{sec:intro}
Content-Based Image Retrieval (CBIR) is a fundamental task in multimedia and computer vision which has undergone a continuous evolution since its early years \cite{smeulders2000content}, moving from the use of engineered features like SIFT to CNNs \cite{zheng2017sift,LI2021675}. It has been applied to many different specialized domains like artworks and cultural heritage \cite{yildirim2018mobile, baldrati2022exploiting}, commerce \cite{Zhan_2021_ICCV, dong2021m5product}, surveillance \cite{ahmad2018object}, nature \cite{ionescu2019imageclef, ionescu2020imageclef}.
In the basic form, the query is composed of only an image, of which features are computed and compared with the ones extracted by a database of images.

\begin{figure}[!tbh]
    \centering
    \includegraphics[width=\linewidth]{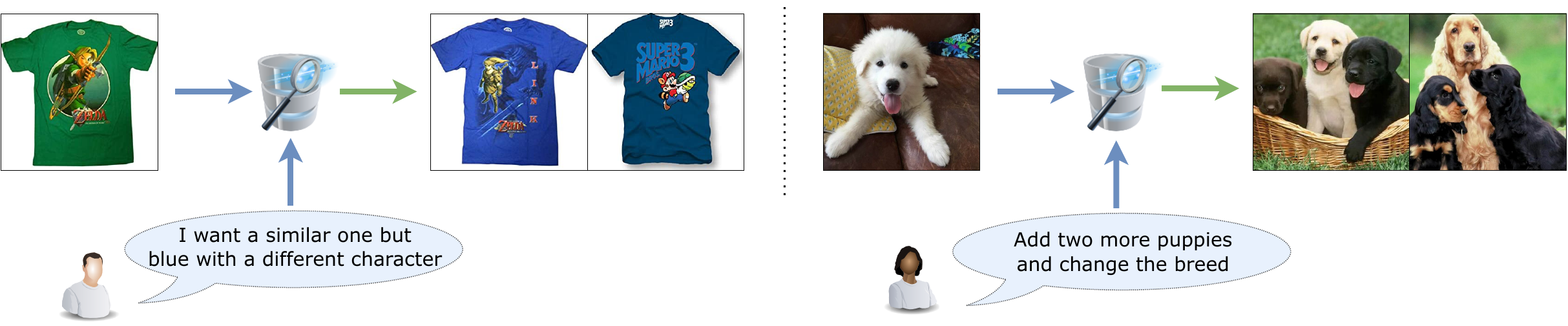}
    \caption{The left portion of the illustration depicts a specific case of composed image retrieval in the fashion domain, where the user imposes constraints on the character attribute of a t-shirt. Meanwhile, the right part showcases an example where the user asks to alter objects and their cardinality within a real-life image.}
    \label{fig:cir-overview}
    \vspace{-2ex}
\end{figure}

We can extend CBIR systems to improve their effectiveness by adding additional information to the query image. For example, interactive image retrieval systems extend CBIR systems by adding some form of user feedback, e.g.~to provide some measure of relevance \cite{banerjee2018relevance}.
In composed image retrieval, the visual query is extended to an image-language pair \cite{liu2021image} where a short textual description, typically expressed in natural language, may request constraints and desired changes or add specifications on some attributes of the retrieved results \cite{Jandial_2022_WACV}. \Cref{fig:cir-overview} illustrates two examples of this task. In both queries, a user selects a reference image and then provides additional requests in the form of text, e.g.~asking to change details, texture, color, or shape features of the reference image. 
Composed image retrieval systems find applications in various domains such as web search, e-commerce, and surveillance. However, developing solutions for this task can be challenging due to the need for incorporating feedback and user intent while addressing the semantic gap between image and text content.

Very recently, researchers proved that deep neural networks combining visual and language modalities like CLIP \cite{radford2021learning}, ALIGN \cite{jia2021scaling}, and the more recent method proposed in \cite{Cheng_2021_CVPR}, trained using an image-caption alignment objective on large-scale internet data, can obtain impressive zero-shot transfer on a myriad of downstream tasks like image classification, text-based image retrieval, and object detection.
%, and video action recognition.

In this work, we show that features obtained from vision and language pretrained (VLP) models -- we employed CLIP-based features -- can be effectively used to implement a composed image retrieval system where user feedback is provided as natural language input to provide additional (or contrasting) requirements concerning those embedded in the visual features of the image used to query the system.
Firstly, we apply the system to the fashion domain, performing experiments on the challenging FashionIQ dataset \cite{wu2021fashion}. Then, to study the generalization capabilities to a broader image domain, we perform experiments on the newly introduced CIRR dataset \cite{liu2021image}. Experiments show that the proposed approach obtains state-of-the-art results on both datasets.

To summarize, we highlight our main contributions as follows:
\begin{itemize}
    \item We propose a novel task-oriented fine-tuning scheme for adapting vision-language models to the composed image retrieval task. The aim of such a task-oriented adaptation scheme is to reduce the mismatch between the large-scale pre-training and the downstream task.

    \item We propose a novel two-stage approach that combines task-oriented fine-tuning with the training of a Combiner network which can perform a fine-grained merging of the multimodal features. This two-stage approach achieves state-of-the-art results on two standard and challenging datasets: FashionIQ and CIRR.

    \item We address the issue of using the CLIP model with images having a high aspect ratio since the CLIP visual encoder can input only square pictures. We propose a novel preprocess pipeline suited for image retrieval tasks that helps to reduce content information loss compared to the standard CLIP preprocess pipeline.
    
    \item To provide further insight into the workings of our proposed system, we perform several qualitative experiments. The first experiment aims to demonstrate how our approach affects the feature distribution in the embedding spaces and the impact of pairwise feature distances on retrieval performance. Additionally, we report visualization experiments utilizing the gradCAM technique \cite{Selvaraju2019gradCAM} to gain a deeper understanding of the image portions that are most significant during retrieval.
\end{itemize}

% \begin{figure*}
%     \centering
%     \includegraphics[width=\linewidth]{images/inference-overview.pdf}
%     \caption{\AB{Ha senso mettere un'immagine in fase di inferenza?}}
%     \label{fig:inference_overview}
% \end{figure*}

\section{Related works}\label{sec:previous}
Traditional CBIR does not use user feedback or its intent to refine results. However, within interactive and composed CBIR, much work has been done to improve retrieval performance by incorporating user's feedback in terms of relevance to the query \cite{718510} or by considering relative \cite{Kovashka_2015} and absolute attributes \cite{8100135, han2017automatic}. The limiting expressiveness of attributes was successively addressed in \cite{vo2019composing, guo2018dialog} by considering purely textual feedback, allowing richer expressiveness. Nonetheless, the performance of the textual model can limit the system in understanding and reacting appropriately. 

\subsection*{Visual and language pre-training}
Models like GPT-2, BERT \cite{devlin2019bert} and GPT-3 \cite{brown2020language} have shown that large amounts of text combined with recent improvements in attention mechanisms enable learning of powerful features that integrate vast knowledge. 
Adding images to the learning process, CLIP \cite{radford2021learning} has very recently shown that it is feasible to perform multimodal zero-shot learning, obtaining remarkable feature generalization of both images and text. 
CLIP is a deep neural network trained to predict the association between text snippets and paired images. Unlike standard vision models trained on specific datasets that are typically good at only one task, this new class of models learns associations between images and natural language supervision that are widely available on the internet.
They are not directly optimized for a benchmark and yet can perform consistently well on different tasks.
CLIP effectiveness is still subject of study \cite{agarwal2021evaluating}, with first applications to art \cite{conde2021clip}, image generation \cite{cimino2021generating} and zero-shot video retrieval \cite{fang2021clip2video}, event classification \cite{li2022clip}, visual commonsense reasoning \cite{wang2022cliptd}. 
Our work builds upon CLIP and further explores its potential in the composed image retrieval task, applying the proposed approach to a specific domain, i.e.~fashion, and also to general images.
ALIGN \cite{jia2021scaling} uses a  dual-encoder architecture to learn the alignment of visual and language representations of image and text pairs using a contrastive loss in a noisy dataset. The extremely large scale of such a dataset, composed of 1 billion pairs, twice the size of the CLIP training dataset, makes up for its noise and leads to state-of-the-art representations even using such a simple learning scheme.
Differently from CLIP and ALIGN, the authors in \cite{Cheng_2021_CVPR} propose a data-efficient contrastive distillation method that learns from a training dataset that is $133\times$ smaller than the one used by CLIP (400 million pairs), using a ResNet50 image encoder and DeCLUTR text encoder.

\subsection*{Composed image retrieval}
In the growing area of image retrieval with user feedback that combines images and text, our work relates to two recently introduced datasets that address the composed image retrieval task: \textit{i)} FashionIQ, a fashion image retrieval with text \cite{wu2021fashion}, and with \textit{ii)} the very recent composed image retrieval of generic images introduced in \cite{liu2021image}. In \cite{Chen_2020_CVPR}, a transformer that can be seamlessly plugged into a CNN to selectively preserve and transform the visual features conditioned on language semantics is presented. Text Image Residual Gating (TIRG) \cite{vo2019composing} combines image and text features using gating and residual features. The authors of \cite{shin2021rtic} leverage skip connections by combining them with graph neural networks, resulting in improved performance. The authors of \cite{Lee_2021_CVPR} employ two different neural network modules to address image style and content.
In \cite{Kim_Yu_Kim_Kim_2021}, the authors present a Correction Network which explicitly models the difference between the reference and target image in the embedding space.  
In \cite{liu2021image}, a new dataset (CIRR) for composed image retrieval on real-world images is proposed, along with a novel transformer-based model that uses rich pre-trained vision-and-language knowledge, called CIRPLANT, to modify visual features conditioned on natural language. CIRPLANT leverages visual-and-language pre-trained models in composed image retrieval: the OSCAR model \cite{li2020oscar} is carefully adapted to the task with promising results.
In \cite{dodds2020modality}, the authors proposed the Modality-Agnostic Attention Fusion (MAAF) model to tackle the composed image retrieval task. The model treats the convolutional spatial image features and learned text embeddings as modality-agnostic tokens and passes them to a Transformer for further processing.
In \cite{Liu-2021-multigrained}, the authors propose a Multi-Grained Fusion (MGF) module which fuses features at different stages.
ComposeAE \cite{Anwaar_2021_WACV} is an autoencoder-based model that learns the composition of image and text features for retrieving images by adopting a deep metric learning (DML) approach instead of fusing them by passing through a few fully connected layers.
CurlingNet, proposed in \cite{yu2020curlingnet}, measures the semantic differential relationships between images concerning a conditioning query text. The main components are two networks: the first one, called the Delivery filter, delivers the source image to the candidate cluster according to a given query in embedding space, while the second one, called the Sweeping filter, checks the attributes highlighted in the query and learns the path from the center of valid target candidates to the target image. 
In \cite{yuan-2021-conversational}, the composed image retrieval task is extended to a multi-turn conversation. The authors proposed a system that utilizes ComposeAE \cite{Anwaar_2021_WACV} to combine image and text at each turn. The combined representation is then fed into a recurrent network, following the turn order, for further processing.
In \cite{Jandial_2022_WACV}, the authors present the SAC (Semantic Attention Composition) framework, which consists of two modules: the Semantic Feature Attention (SFA) module finds the salient regions of the image w.r.t.~the text, and then the Semantic Feature Modification (SFM) module determines how to change the relevant parts of the image compositing coarse and fine salient image features computed by SFA with text embeddings.

The proposed method starts with the hypothesis of having a unified embedding space for images and text achieved through the Vision-Language model CLIP. In the first stage, we fine-tune both CLIP encoders to adapt them to the composed image retrieval task. Next, using the task-adapted embedding spaces, we train a Combiner network to merge the multimodal features. In contrast to fashion-oriented approaches like~\cite{Chen_2020_CVPR, Lee_2021_CVPR}, our method does not rely on spatial features. Instead, we argue that when considering images of a broader domain, the semantics hold greater significance than local visual aspects.

% Our method differs from these works by a few factors.
% It relies on a large-scale pretrained Vision-Language model and it breaks up the CLIP common embedding space with the goal of learning an additive transformation in the image space with the guidance of textual information. 
% Moreover, our approach does not use any kind of spatial information. Instead, in \cite{vo2019composing,Lee_2021_CVPR} features extracted from the backbone are 3-dimensional and the composition takes care of spatial information, in \cite{Chen_2020_CVPR} the features are extracted at different convolutional layers from the ResNet-50 backbone. In \cite{Kim_Yu_Kim_Kim_2021} the authors divided the image and the sentence into a set of localized components assigning a representation module, denoted as \textit{experts}, to each of them. More similar to our work is \cite{shin2021rtic}, which trains a combiner directly on the flattened image and text features that, differently from our work, are obtained from different embeddings. 

\section{The proposed method}\label{sec:method}
The proposed approach addresses the multimodal task of composed image retrieval. The input query consists of a reference image $I_q$ (e.g., an image of a black shirt with a cartoon lion) and a relative caption $T_q$ that includes a descriptive request from the user about the image (e.g., "has dog print and is dark grey color"). The goal is to retrieve target images that satisfy similarity constraints imposed by both the input components (e.g., an image of a dark grey shirt with a dog print, as shown in \cref{fig:training}). For a successful retrieval, the system should understand the semantics of the image and the meaning of the text, integrate the multi-domain information, and then use the fused representation to retrieve the relevant images.

In contrast to previous works like \cite{Kim_Yu_Kim_Kim_2021, shin2021rtic, Chen_2020_CVPR, Lee_2021_CVPR} that build from different image and textual models, we start from the hypothesis of having a unified embedding of images and text, obtained through using the CLIP model~\cite{radford2021learning}.
CLIP is a vision-language model trained to align images and their corresponding text captions in a unified embedding space. It consists of an image encoder $\psi_{I}$ and a text encoder $\psi_{T}$. Given an image $I$, the image encoder extracts a feature representation $\psi_{I}(I) \in \mathbb{R}^{d}$, where $d$ is the size of the CLIP embedding space. Similarly, for a given text caption $T$, the text encoder extracts a feature representation $\psi_{T}(T) \in \mathbb{R}^{d}$. CLIP learns to map similar concepts expressed in images and text to similar feature representations.
For instance, given the image of a cat $I_c$ and the text $T_c$ \textit{``a photo of a cat"}, the way CLIP is trained should guarantee that $\psi_{I}(I_c) \approx \psi_{T}(T_c)$.

We argue that, even though having a unified embedding space is a good starting point, it is not exactly what we need in the task we are considering. 
In composed image retrieval, the goal is to move from the reference to the target image point in the image embedding space with the aid of textual information. Hence, instead of utilizing a unified image-text embedding space, our approach involves creating two separate embedding spaces that can be combined through a sum operation.
Formally, given an image of a black dress $I_x$ and the corresponding text $T_y$ ("is blue"). Let $I_z$ represent the image of a blue dress. Our aim is to shape the embedding spaces such that $\psi_{I}(I_x) + \psi_{T}(T_y) \approx \psi_{I}(I_z)$. When this equation is satisfied, we can affirm that the textual embedding space exhibits strong "additivity properties" in relation to the image space, or equivalently, the embedding spaces are additive. Ideally, the embeddings of the relative caption should correspond to the displacement vector from the query image to the target image features, i.e. $\psi_{T}(T_y) \approx \psi_{I}(I_z) - \psi_{I}(I_x)$.

We propose a two-stage approach to address the task of composed image retrieval by taking full advantage of the capabilities of CLIP's features.
In the first stage, we tackle the objective mismatch between the large-scale pretraining of CLIP and the downstream task: we propose a novel fine-tuning scheme tailored to improving the additivity properties of the embedding spaces. In the second stage, starting from the task-oriented features, we train from scratch a Combiner neural network that learns to perform a fine-grained combination of image-text features. Although we train the Combiner network from scratch, we design its structure to take full advantage of the first stage of training (see \cref{sec:combiner_training} for more details).
During both stages the training is performed using triplets ($I_q$, $T_q$, $I_t$), where $q = (I_q, T_q)$ is the query and $I_t$ is the target image that we aim to retrieve given $q$. 

At inference time, given a query $(I_q, T_q)$, we utilize the fine-tuned CLIP encoders and the trained Combiner network to generate the combined features. Subsequently, following the standard image-to-image retrieval approach, we compute the cosine distances between the combined features and the database of index image features. The results are then sorted based on their similarity.

\begin{figure*}[t]
  \includegraphics[width=0.95\textwidth]{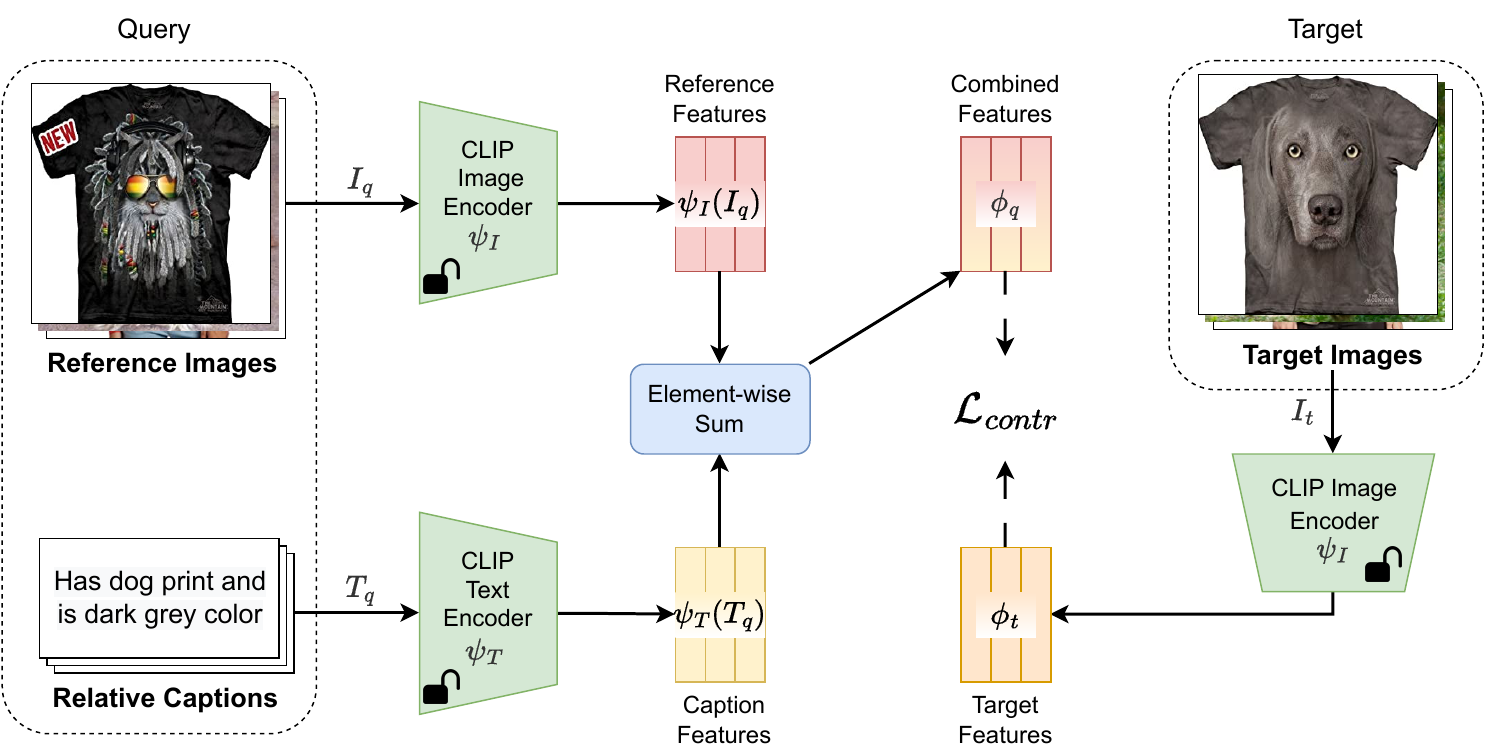}
  \caption{First stage of training. In this stage, we perform a task-oriented fine-tuning of CLIP encoders to reduce the mismatch between the large-scale pre-training and the downstream task. We start by extracting the 
 image-text query features and combining them through an element-wise sum. We then employ a contrastive loss to minimize the distance between combined features and target image features in the same triplet and maximize the distance from the other images in the batch. We update the weights of both CLIP encoders.}
  \label{fig:fine-tuning}
\vspace{-2.5ex}
\end{figure*}

\subsection{Task-oriented fine-tuning}
In this stage, we adapt both CLIP's encoders to composed image retrieval reducing the mismatch between the large-scale pre-training and the downstream task. 
Given a query consisting of a reference image $I_q$ and a relative caption $T_q$, we extract their feature representations using the CLIP image encoder $\psi_{I}$ and text encoder $\psi_{T}$ respectively. This results in $\psi_{I}(I_q) \in \mathbb{R}^{d}$ and $\psi_{T}(T_q) \in \mathbb{R}^{d}$, where $d$ denotes the size of the CLIP embedding space.
To combine the query features, we perform an element-wise sum, resulting in $\phi_q = \psi_{I}(I_q) + \psi_{T}(T_q)$.

Our objective is to minimize the distance between the query combined features $\phi_q$ and the target image features $\phi_t = \psi_{I}(I_t)$ belonging to the same triplet and, at the same time, maximize the distance from the other target images in the same batch.
To this end, following \cite{vo2019composing, shin2021rtic, Lee_2021_CVPR}, we employ a batch-based contrastive loss:
\begin{equation}
    \mathcal{L}_{contr} = \frac{1}{B} \sum^{B}_{i=1} -\text{log} \frac{\text{exp\{} \tau * \kappa(\phi_q^i, \phi_t^i)\}}{\sum^B_{j=1} \text{exp\{} \tau * \kappa(\phi_q^i, \phi_t^j)\}}
    \label{eq:loss}
\end{equation}
Here, $\kappa(\cdot)$ denotes the cosine similarity, $\tau$ is a temperature parameter that controls the range of the logits, and $B$ is the number of images in a batch. We update the weights of both CLIP encoders.
We use this loss because, being a batch-wise contrastive loss, it does not require the definition of a sampling strategy: it considers all negative samples in a mini-batch.
\Cref{fig:fine-tuning} shows an overview of the task-oriented fine-tuning stage.

% \Cref{fig:fine-tuning} shows an overview of the task-oriented fine-tuning stage. In this stage, we adapt both CLIP's encoders to the task performing a fine-tuning that breaks up the symmetry between the encoders.
% Initially, the input images and texts are encoded using their respective encoders into the features space. Then the query features are combined using an element-wise sum followed by an $L_2$-normalization. Finally, with the aid of a contrastive loss, which compares the combined features and the target features in a batch, the weights of both CLIP encoders are updated.

Using the element-wise sum as the combination of query features goes in the direction of making CLIP's embedding spaces more additive. Consequently, similar concepts expressed in text and images no longer share similar features. Instead, the textual features serve as displacement vectors from the query to the target in the image space.
From a high-level perspective, we notice that, in composed image retrieval, the image and the text do not play the same role. The task is not symmetric with respect to the input: we start from an image, and we would like to retrieve another image using textual guidance.
For this reason, the break up of the unified embedding space is not an undesirable side-effect.

We will denote the fine-tuned image encoder and text encoder as $\overline{\psi_{I}}$ and $\overline{\psi_{T}}$, respectively.

\subsection{Combiner training}\label{sec:combiner_training}
\begin{figure*}[tb]
 \includegraphics[width=0.95\textwidth]{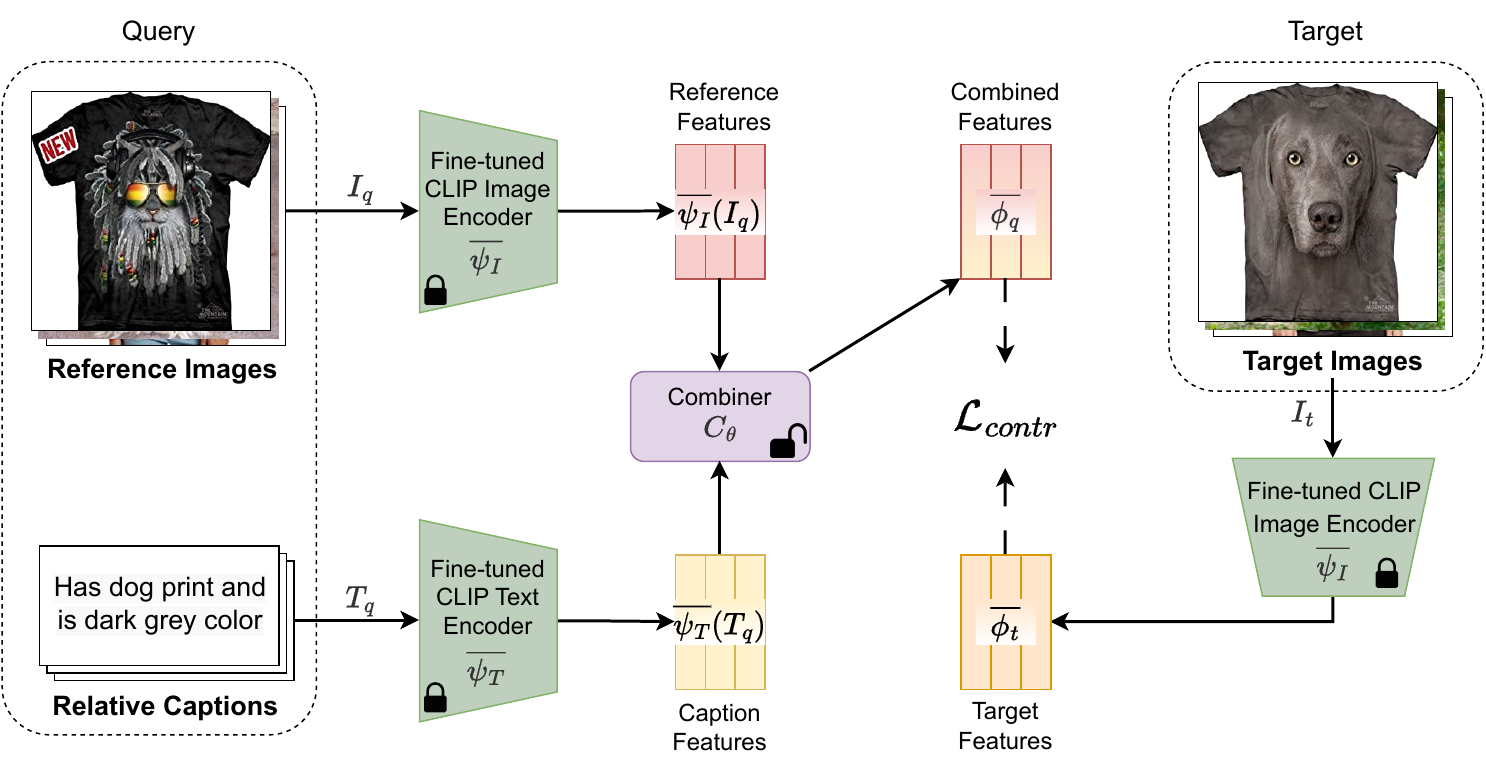}
  \caption{Second stage of training. In this stage, we train from scratch a Combiner network that learns to fuse the multimodal features extracted with CLIP encoders. We start by extracting the image-text query features using the fine-tuned encoders, and we combine them using the Combiner network. We then employ a contrastive loss to minimize the distance between combined features and target image features in the same triplet and maximize the distance from the other images in the batch. We keep both CLIP encoders frozen while we only update the weights of the Combiner network.
  At inference time the fine-tuned encoders and the trained Combiner are used to produce an effective representation used to query the database.}
  \label{fig:training}
\end{figure*}

During the training of the Combiner network, we follow the same general framework as in the previous stage. However, this time we train from scratch the Combiner network instead of updating the weights of the CLIP encoders.
In contrast to the first stage, we use the Combiner network $C_{\theta}$ to combine the query features. Specifically, the combined features are obtained as $\overline{\phi_q} = C_{\theta}(\overline{\psi_{I}}(I_q), \overline{\psi_{T}}(T_q))$.
We optimize the Combiner network by utilizing the $\mathcal{L}_{contr}$ loss described in \cref{eq:loss} with $\overline{\phi_q}$ and $\overline{\phi_t} = \overline{\psi_{T}}(I_t)$ as inputs. \Cref{fig:training} depicts a visual overview of the Combiner network training stage.
By employing the contrastive loss, we train the Combiner $C_{\theta}$ to produce features as close as possible to the target features and as far away as possible from all other image features.

\begin{figure*}[tb]
    \centering
    \includegraphics[width=\linewidth]{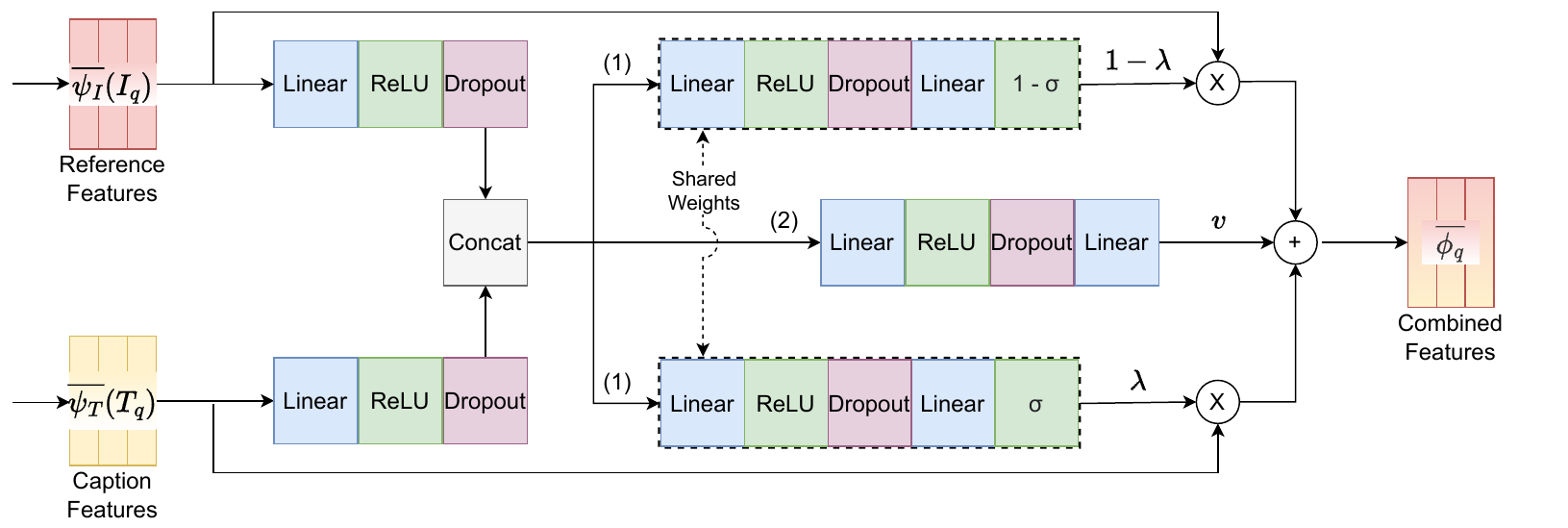}
    \caption{Architecture of the Combiner network $C_{\theta}$. It takes as input the multimodal query features and outputs a unified representation. $\sigma$ represents the sigmoid function. We denote the outputs of the first branch (1) as $\lambda$ and $1 -\lambda$, while the output of the second branch (2) as $v$. The combined features are $\overline{\phi_q} = (1 - \lambda)* \overline{\psi_{I}}(I_q) + \lambda * \overline{\psi_{T}}(T_q) + v$}
    \label{fig:combiner}
\end{figure*}

The Combiner network, depicted in \cref{fig:combiner}, is designed to take full advantage of the first stage of training and the increased additivity properties of the adapted embedding spaces. The idea is to learn the residual of a convex combination of the image-text query features.
We begin by projecting the text and image features through a linear transformation followed by a ReLU function. The resulting projected features are then concatenated and passed to two separate branches. The first branch, labeled as (1) in \cref{fig:combiner}, is responsible for computing the coefficients of a convex combination between the image and text features.
To compute these coefficients, we feed the concatenated features into a linear layer, followed by the ReLU function, another linear layer, and the sigmoid function. The sigmoid output provides the coefficients needed for the query image-text convex combination.
The second branch, labeled as (2), outputs the mixture contribution of the image and text features. The structure of this branch is the same as the first branch, except it does not include the final sigmoid function.
Finally, we sum the convex combination of the query features and the learned image-text mixture. To reduce overfitting, we apply dropout to each layer.

By denoting the outputs of the first branch (1) as $\lambda$ and $1 -\lambda$, and the output of the second branch (2) as $v$, we can express the combined features as $\overline{\phi_q} = (1 - \lambda)* \overline{\psi_{I}}(I_q) + \lambda * \overline{\psi_{T}}(T_q) + v$. Notably, the convex combination $(1 - \lambda)* \overline{\psi_{I}}(I_q) + \lambda * \overline{\psi_{T}}(T_q)$ is a generalization of the element-wise sum of the query features. Consequently, as the embedding spaces exhibit stronger additivity properties, the Combiner's effectiveness in its task is enhanced. We intentionally design the Combiner to capitalize on the task adaptation achieved during the first stage of training.

\subsection{Preprocess Pipeline}\label{sec:preprocess}
\begin{figure}[tb]
       \centering
    \begin{subfigure}{0.45\linewidth}
    \includegraphics[width=\linewidth]{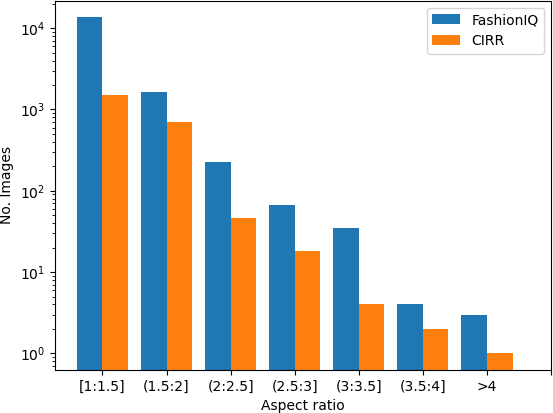}
    \caption{Dataset comparison}
    \end{subfigure}\hfill
 \begin{subfigure}{0.45\linewidth}
    \includegraphics[width=\linewidth]{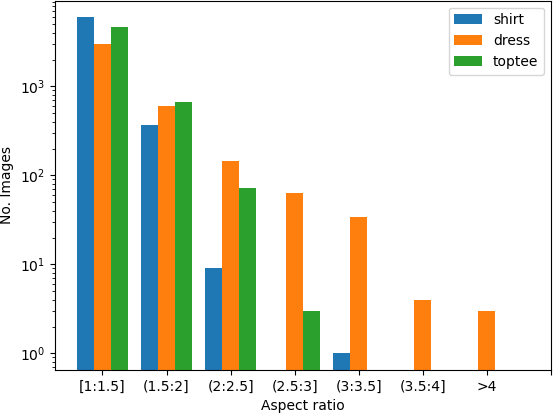}
    \caption{FashionIQ categories}
    \end{subfigure}
     \vspace{-1.8ex}
    \caption{Histogram of image aspect ratios in FashionIQ and CIRR datasets (a) and the three categories of FashionIQ (b). The x-axis represents the aspect ratio defined as $max(width, height) / min(width, height)$ while the y-axis represents the number of images (in logarithmic scale). The width of each bin is 0.5, and the first bin starts at 1. More than half of the dataset's images are skewed and have at least a 1.5 aspect ratio. In the FashionIQ dataset, the issue is evident in the Dress category.}
    \label{fig:hist_plot}
    \vspace{-3ex}
\end{figure}

The standard preprocess pipeline of CLIP is mainly composed of two steps: a resize operation where the smaller side of the image matches the CLIP input dimension $input\_dim$ followed by a center crop operation which results in a square patch  $input\_dim\times input\_dim$ output.
Subsequently, as the ratio between the largest and the smaller side increases, the area of the image lost after the preprocess increases. From now on, we will say that an image has a high aspect ratio when it is far from having a square shape.
In \cref{fig:hist_plot} is shown how, in the datasets we consider (detailed in \cref{sec:experiments}), the number of images with a high aspect ratio is not negligible. As can be seen, this is especially true for the FashionIQ \emph{dress} category and the CIRR dataset.

\begin{figure}[t]
\centering
\begin{minipage}[t!]{0.50\textwidth}
\begin{algorithm}[H]
\centering
   \caption{Python-style pseudocode of the proposed preprocess pipeline.}
   \label{alg:preprocess}
   
    \definecolor{codeblue}{rgb}{0.25,0.5,0.5}
    \lstset{
      basicstyle=\fontsize{7.2pt}{7.2pt}\ttfamily\bfseries,
      commentstyle=\fontsize{7.2pt}{7.2pt}\color{codeblue},
      keywordstyle=\fontsize{7.2pt}{7.2pt},
    }
\begin{lstlisting}[language=python]
# in_image: input image to be preprocessed
# target_ratio: target aspect ratio
# dim: CLIP image encoder input dimension

def preprocess(in_image, target_ratio, dim):
    w, h = in_image.size
    aspect_ratio = max(w, h) / min(w, h)

    # pad the image only if the aspect ratio 
    # is above a fixed target
    if aspect_ratio < target_ratio:
        out_image =  in_image
    else:
        # zero-pad the image to bring its aspect
        # ratio to target ratio
        scaled_max_wh = max(w, h) / target_ratio
        hp = max((scaled_max_wh - w) // 2, 0)
        vp = max((scaled_max_wh - h) // 2, 0)
        padding = (hp, vp, hp, vp)
        out_image = pad(in_image, padding, 0)

    # Resize and center crop the image
    out_image = resize(out_image, dim)
    out_image = center_crop(out_image, dim)                               
return out_image
\end{lstlisting}
\end{algorithm}
\end{minipage}
\hfill
\begin{minipage}[c]{0.46\textwidth}
        \centering
        \includegraphics[width=\linewidth]{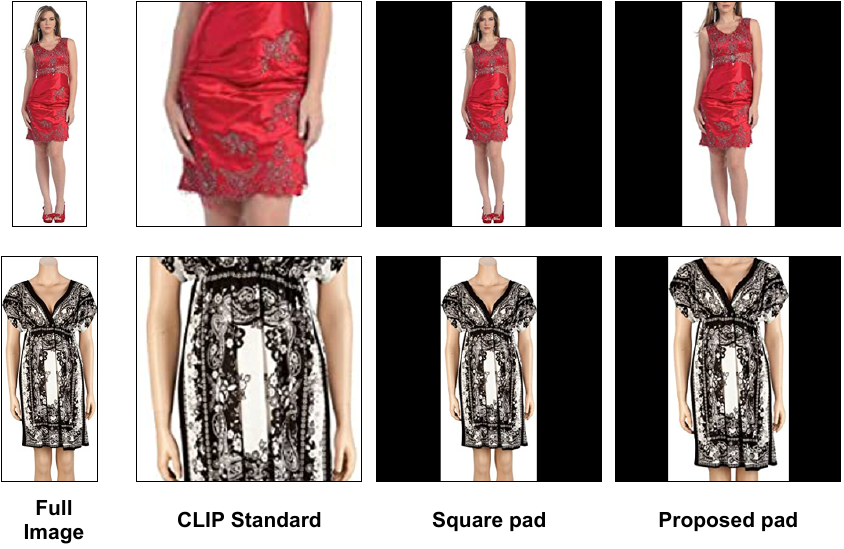}
        \caption{Comparison among different preprocesses pipelines. The proposed padding method results in images that contain more details than square padding and provides a better overview than the standard CLIP padding.}
        \label{fig:pre_comp}
    \end{minipage}
    \vspace{-3ex}
\end{figure}

One way to address the loss of information due to the center crop operation is to apply zero-padding to match the smaller side to the larger side, effectively squaring the image. 
Although this approach eliminates the loss of content information, it also reduces the resolution of the useful portion of the image since the CLIP image encoder input dimension cannot change.
Thus, we develop a preprocessing pipeline that seeks to balance the two approaches discussed above. Specifically, we apply padding to an image only if its aspect ratio exceeds a predefined target ratio. Additionally, instead of squaring the image, we adjust its aspect ratio to match the target ratio when padding is applied.
The pseudocode for the proposed preprocess pipeline is shown in Algorithm \ref{alg:preprocess}.

\Cref{fig:pre_comp} presents the preprocess pipelines as mentioned earlier. It is evident that when the ratio between the larger and smaller sides deviates significantly from one, the standard CLIP preprocess removes a substantial portion of the image, which considerably hampers the retrieval process. Although the visual disparities between the square pad and the proposed pad (with a target ratio of 1.25) approaches are not substantial, we will demonstrate that the model benefits from having such an increased usable portion in the images during retrieval.

% \begin{figure}[tbh]
%     \centering
%     \includegraphics[width=0.6\linewidth]{images/preprocess_comparison.pdf}
%     \caption{Comparison among different preprocesses pipelines. The proposed padding method results in images that contain more details than square padding and provides a better overview than the standard CLIP padding.}
%     \label{fig:pre_comp}
% \end{figure}

\section{Experimental results}\label{sec:experiments}
\subsection{Implementation details}
We perform the experiments using two CLIP models of different sizes. The smallest one relies on a modified ResNet-50 (RN-50) \cite{he2016deep} architecture. It takes input images of $224\times224$, and the size of its embedding space is $d=1024$. The biggest one, denoted as RN-50x4, follows the EfficientNet-style model scaling and uses approximately $4\times$ the computation of RN-50. It takes input images of $288\times288$, and the size of its embedding space is $d=640$.

In the Combiner network (\Cref{fig:combiner}), the first two linear layers before the concatenation have input-output dimensionality equal to $(d, 4d)$.
After the concatenation in both branches, we have two linear layers. In the first branch (1), the first linear layer has input-output dimensionality of $(4d, 8d)$, and the second one $(8d, 1)$.
In the other branch, the first linear layer has input-output dimensionality of $(4d, 8d)$ while the second one $(8d, d)$.
Following the standard practice, we set the dropout rate to 0.5.
During retrieval, we normalize both the combined and index set features to have a unit $L_2$-norm.

Following the original CLIP training strategy, in the fine-tuning stage, we employed AdamW optimizer \cite{loshchilov2018decoupled} with a learning rate of $2e-6$ and a weight decay coefficient of $1e-2$. 
Due to GPU memory constraints, we set the batch size to 512 for fine-tuning the RN-50-based CLIP model and 192 for fine-tuning the RN-50x4-based model.
We kept the batch normalization layer frozen.
We fine-tuned the CLIP encoders for a maximum of 150 epochs.
During the training of the Combiner network, we keep both fine-tuned CLIP encoders frozen and only train the Combiner function. We set the learning rate to $2e-5$ and train the model for a maximum of 300 epochs. We set the batch size to 4096 when using both backbones.
We used the PyTorch library throughout the experiments. We set the target ratio in the preprocessing pipeline to 1.25. 
Following the approach described in \cite{radford2021learning}, we set the parameter $\tau$ in \cref{eq:loss} to 100. This value ensures that the logits have a sufficient dynamic range.
To mitigate overfitting, we adopt an early stopping strategy. We use mixed-precision training \cite{micikevicius2018mixed} to save memory and speed up the training in both stages. We employ gradient checkpointing \cite{chen2016training} to further reduce memory usage.

We conduct all experiments on a single NVIDIA Titan RTX (24GB) GPU. The first stage of training requires approximately 4 hours for the RN-50x4 model and 2 hours for the RN-50 model. The training of the Combiner network takes less than an hour for both models.

\subsection{Datasets and metrics}\label{sec:dataset}

\subsubsection{FashionIQ}
FashionIQ \cite{wu2021fashion} is composed of 77,684 fashion images crawled from the web and split into the train, validation, and test sets, divided into three different categories: \textit{Dress}, \textit{Toptee} and \textit{Shirt}. Among the 46,609 training images, there are 18,000 training triplets made of a reference image, a pair of relative captions, and a target image. The captions describe properties to modify in the reference image to match the target image. 
The validation and test sets consist of 15,537 and 15,538 images, respectively, with 6,017 and 6,119 triplets. 
%Additionally, there are 29,728 images annotated with training attributes available for use in the training process, but we chose not to utilize them.

We follow the standard experimental setting as in \cite{Lee_2021_CVPR, Kim_Yu_Kim_Kim_2021}. We employ the average recall at rank K (Recall@K) as an evaluation metric, namely Recall@10 (R@10) and Recall@50 (R@50). Note that for each triplet, there is only a positive index image. Hence, each query has R@K zero or one.
All results are on the validation set since, at the time of writing, test set ground-truth labels have not been released yet.

\subsubsection{CIRR} The authors of \cite{liu2021image} designed the CIRR dataset to address two common problems encountered in composed image retrieval datasets, such as FashionIQ. These problems are the lack of sufficient visual complexity caused by the restricted image domain and the numerous false negatives due to the unfeasibility of extensively labeling target images for each (reference, text) pair. As a result, some images in the dataset that correspond to valid matches for a query are not labeled as valid targets.
% That is, images $I \in D$ that are valid matches for the query but not labeled as the ground-truth target $I_T$. Indeed, all images in $D/{I_R, I_T}$ are considered negatives. To circumvent this shortcoming, existing works choose to evalutate models with Recall@$K$ and set $K$ to larger values (\textit{e.g.} 10, 50 \cite{wu2021fashion}), thus accounting for the presence of false-negatives. However, the issue persists during training. Moreover, by setting larger K values, these methods are essentially trading in their ability for learning detailed text-image modifications.
CIRR (Compose Image Retrieval on Real-life images) dataset consists of 21,552 real-life images taken from the popular natural language reasoning $NLVR^2$ dataset \cite{suhr2019corpus}. It has the same structure as the FashionIQ dataset and contains 36,554 triplets randomly assigned in 80\% for training, 10\% for validation, and 10\% for the test.
The dataset images are grouped in multiple subsets of six semantically and visually similar images. To have negative images with high visual similarity the relative captions are collected describing the differences between two images in the same subset. 

The standard evaluation protocol proposed by the authors of the dataset is to report the recall at rank K (Recall@K) at four different ranks (1, 5, 10, 50). Moreover, thanks to the unique design of the CIRR dataset, it is also possible to report the Recall$_\text{Subset}$ metric that considers only the images in the subset of the query. This \emph{subset} metric has two main benefits: it is not affected by false-negative samples and, thanks to negative samples with high visual similarity, it captures fine-grained image-text modifications.
The reference metrics are the R@5 which accounts for possible false negatives in the entire corpus, and the R$_\text{Subset}$@1, which better illustrates the fine-grained reasoning abilities.

\subsection{Task-oriented fine-tuning effects} \label{sec:fine-tuning}
In this section, we present a set of experiments that illustrate how the task-oriented fine-tuning of CLIP encoders and their increased additivity properties contribute to easing the task of the Combiner network and help to improve retrieval performance.
For each dataset, we compare the performance varying the combining function and the modality of the CLIP fine-tuning.
Throughout all the experiments, we use the RN-50 CLIP model. For each fine-tuning modality, we train from scratch a different Combiner network.
We report the results in Table \ref{tab:fashioniq-finetune} for the FashionIQ dataset and in Table \ref{tab:cirr-finetune} for the CIRR dataset. 

Notably, the element-wise sum of out-of-the-box CLIP features achieves impressive results without domain or task-specific training on both datasets. This performance is intriguing as it demonstrates that the CLIP image-text common embedding space exhibits good additivity properties, even though its training objective does not explicitly optimize for this aspect.
Fine-tuning only the CLIP image encoder brings an interesting performance boost compared to the out-of-the-box CLIP features. This improvement is expected when employing the element-wise sum as the combining function, given that the out-of-the-box CLIP features lack domain or task-specific training. However, the most promising improvement occurs when utilizing the trained Combiner network.
The text encoder fine-tuning achieves slightly better performance than image encoder fine-tuning. We can notice that on the FashionIQ dataset, the improvement over the image encoder fine-tuning remains constant when using either the element-wise sum or the Combiner network as a combining function. 
However, on the CIRR dataset, the situation differs. When comparing with the performance of the image encoder fine-tuning, using the element-wise sum to combine the query features results in comparable global metrics, but significantly improved fine-grained \textit{subset} metrics. In contrast, when utilizing the Combiner network, we observe a reduction in the gaps within the \textit{subset} metrics, while achieving a greater improvement in the global metrics.
We achieve the best results on both datasets when we fine-tune both encoders. The element-wise sum of the fine-tuned features outperforms the performance of the out-of-the-box features combined with the trained Combiner network by a significant margin. Moreover, when we combine the query features with the Combiner network, the performances further improve. It is worth highlighting that when utilizing the Combiner as a combining function, the improvement achieved by fine-tuning both encoders over the out-of-the-box CLIP features is the arithmetic sum of the improvements obtained by fine-tuning either the image or the text encoder.

\begin{table}[tb]
\centering
% \resizebox{1.0\columnwidth}{!}{%
\begin{tabular}{ccc  cc cc cc  cc}
\toprule

&&&\multicolumn{2}{c}{\textbf{Shirt}} & \multicolumn{2}{c}{\textbf{Dress}} & \multicolumn{2}{c}{\textbf{Toptee}} & \multicolumn{2}{c}{\textbf{Average}}\\
\cmidrule(lr){4-5}
  \cmidrule(lr){6-7}
  \cmidrule(lr){8-9}
  \cmidrule(lr){10-11}
\textbf{CF} & \textbf{IFT} & \textbf{TFT} & $R@10$ & $R@50$ &  $R@10$  & $R@50$ &  $R@10$  & $R@50$ &  $R@10$  & $R@50$  \\
\midrule
\multirow{4}{*}{Sum} & \xmark & \xmark & 19.53 & 35.57 & 17.70 & 36.29 & 21.88 & 42.93 & 19.70 & 38.26\\ 
&\cmark& \xmark & 30.08 & 52.94 & 29.10 & 52.01 & 34.42 & 57.62 & 31.20 & 54.19\\
&\xmark& \cmark & 32.29 & 53.73 & 27.76 & 52.31 & 35.14 & 60.12 & 31.73 & 55.39\\
&\cmark& \cmark & \underline{38.67} & \underline{59.42} & \underline{35.99} & \underline{62.22} & \underline{43.35} & \underline{67.52} & \underline{39.34} & \underline{63.05}\\ \midrule[.02em]

\multirow{4}{*}{Combiner} & \xmark & \xmark & 31.85 & 52.50 & 27.22 & 50.62 & 33.81 & 57.57 & 30.96 & 53.56\\ %111
&\cmark& \xmark & 34.30 & 55.79 & 32.47 & 55.18 & 38.45 & 62.36 & 35.07 & 57.78\\ %rifare
&\xmark& \cmark & 35.87 & 57.21 & 31.43 & 54.98 & 38.20 & 63.22 & 35.16 & 58.47\\ % 96
&\cmark& \cmark & \textbf{39.87} & \textbf{60.84} & \textbf{37.67} & \textbf{63.16} & \textbf{44.88} & \textbf{68.59} & \textbf{40.80} & \textbf{64.20} \\ \bottomrule %39
\end{tabular}
% }
\caption{Recall at K on the FashionIQ validation set while varying the combining function and the modality of CLIP fine-tuning. We denote \textbf{IFT} (image encoder fine-tuning) and \textbf{TFT} (text encoder fine-tuning) to represent whether the image encoder or the text encoder is fine-tuned in the first stage. \textbf{CF} (combining function) indicates the function used to combine the query features. We highlight the best scores in bold and underline the second-best scores.}
\vspace{-3ex}
\label{tab:fashioniq-finetune}
\end{table}

\begin{table}[tb]
\centering
% \resizebox{1.00\columnwidth}{!}{%
\begin{tabular}{ccc  cccc ccc}
  \toprule

&&& \multicolumn{4}{c}{\textbf{Recall@K}} & \multicolumn{3}{c}{\textbf{R$_\text{subset}$@K}} \\
\cmidrule(lr){4-7}
\cmidrule(lr){8-10}
\textbf{CF} & \textbf{IFT} & \textbf{TFT}  & $K=1$ & $K=5$ &  $K=10$ & $K=50$ &  $K=1$ & $K=2$ &  $K=3$ \\
\midrule
\multirow{4}{*}{Sum} & \xmark & \xmark & 21.38 & 50.85 & 64.00 & 87.23 & 54.48 & 76.01 & 87.16 \\
&\cmark& \xmark & 31.67 & 66.08 & 79.36 & 95.38 & 58.12 & 78.42 & 89.78 \\
&\xmark& \cmark & 32.72 & 66.63 & 79.22 & 94.86 & 67.21 & 86.00 & 93.81 \\
&\cmark& \cmark & \underline{40.97} & \underline{74.70} & \underline{85.51} & \underline{96.94} & \underline{68.81} & \underline{86.96} & 93.90\\ \midrule[.02em]

\multirow{4}{*}{Combiner} & \xmark & \xmark & 31.26 & 64.79 & 77.71 & 95.31 & 61.56 & 81.08 & 91.12 \\ %75
&\cmark& \xmark & 34.01 & 69.07 & 81.77 & 95.72 & 62.78 & 81.80 & 91.41 \\ %45
&\xmark& \cmark & 36.86 & 71.32 & 82.32 & 96.24 & 68.28 & 86.51 & \underline{94.14} \\ %54
&\cmark& \cmark & \textbf{42.05} & \textbf{76.13} & \textbf{86.51} & \textbf{97.49} & \textbf{70.15} & \textbf{87.18} & \textbf{94.40}\\ \bottomrule % 33
\end{tabular}
% }
\caption{Recall at K on the CIRR validation set while varying the combining function and the modality of CLIP fine-tuning. We denote \textbf{IFT} (image encoder fine-tuning) and \textbf{TFT} (text encoder fine-tuning) to represent whether the image encoder or the text encoder is fine-tuned in the first stage. \textbf{CF} (combining function) indicates the function used to combine the query features. We highlight the best scores in bold and underline the second-best scores.}
\label{tab:cirr-finetune}
\vspace{-4ex}
\end{table}
Given this last observation and all the other results, we formulate the hypothesis that the fine-tuning of the image and the text encoder learn different and complementary information that improves performances differently.
We conjecture that the fine-tuning of the image encoder adapts the image manifold to the domain of the data (e.g., the fashion domain for the FashionIQ dataset). On the contrary, the fine-tuning of the text-encoder adapts the text embedding space to the task of composed image retrieval by transforming textual features into displacement vectors within the image embedding space.
%by making the text manifold contains displacement vectors in the image space. \AB{Forse frase da cambiare?}
In support of this conjecture, we highlight the difference in performances between the global metrics and \textit{subset} metrics on the CIRR dataset when comparing the image and the text encoder fine-tuning using the element-wise sum as a combining function (second and third row in \Cref{tab:cirr-finetune}). 
We note that in the global metrics, where the domain of the images is diverse, the performance differences between the two experiments approach zero. Conversely, in the \textit{subset} metrics, where the visual differences among the images are low, the image fine-tuning is not capable of capturing the fine-grained differences making the textual information more discriminative and thus making the fine-tuning of the text encoder perform better.
The experiments described in \cref{sec:feat_study} provide additional confirmation of our intuition.

\subsection{Combiner ablation study}\label{sec:ablation}
In this section, we present a set of experiments with ablations and variations of the proposed Combiner network. We perform all the experiments using the fine-tuned RN-50 CLIP model. We train all the Combiner networks using a batch size of 4096 and a learning rate of $2e-5$.

Given the proposed Combiner network illustrated in \cref{fig:combiner}, we denote the outputs of the first branch (1) as $\lambda$ and $1 -\lambda$, while the output of the second branch (2) as $v$.
The output features of the proposed Combiner are: $\overline{\phi_q} = (1 - \lambda)* \overline{\psi_{I}}(I_q) + \lambda * \overline{\psi_{T}}(T_q) + v$.

To evaluate each component of the proposed design, we tested the following variations:

\begin{itemize}
    \item \textbf{Element-wise sum}: fine-tuned image and text features are summed: $\overline{\phi_q} = \overline{\psi_{I}}(I_q) + \overline{\psi_{T}}(T_q)$
    \item \textbf{Convex combination}: only convex combination of image and text features, i.e. the model without the mixture contribution of text and image: $\overline{\phi_q} = (1 - \lambda)* \overline{\psi_{I}}(I_q) + \lambda * \overline{\psi_{T}}(T_q)$
    \item \textbf{W/o convex combination}: only the mixture contribution of text and image, i.e the model without the convex combination of text and image features: $\overline{\phi_q} = v$
    %\item \textbf{Linear after skip}: the proposed model with an additional linear layer before the $L_2$ normalization;
    \item \textbf{Static skip}: the convex coefficients are statically set to 0.5: $\overline{\phi_q} = 0.5 * \overline{\psi_{I}}(I_q) + 0.5 * \overline{\psi_{T}}(T_q) + v$
    %\item \textbf{No Dropout}: Combiner network without dropout layers;
    %\item \textbf{No ReLU}: Combiner network without ReLU activations;
    \item \textbf{Proposed Combiner}: the Combiner architecture illustrated in \cref{fig:combiner}.
\end{itemize}

\begin{table}[tb]
\centering
\begin{tabular}{lcc  cc cc cc  cc}
\toprule

&\multicolumn{2}{c}{\textbf{Shirt}} & \multicolumn{2}{c}{\textbf{Dress}} & \multicolumn{2}{c}{\textbf{Toptee}} & \multicolumn{2}{c}{\textbf{Average}}\\
  \cmidrule(lr){2-3}
\cmidrule(lr){4-5}
  \cmidrule(lr){6-7}
  \cmidrule(lr){8-9}
\textbf{Model} & $R@10$ & $R@50$ &  $R@10$  & $R@50$ &  $R@10$  & $R@50$ &  $R@10$  & $R@50$  \\
\midrule
Element-wise sum  & 38.67 & 59.42 & 35.99 & 62.22 & 43.35 & 67.52 & 39.34 & 63.05\\ 
Convex combination & \underline{39.45} & 60.16 & 36.44 & 62.57 & 44.05 & 67.87 & 39.98 & 63.53 \\  % 24
W/o convex combination & 31.40 & 55.64 & 35.94 & 61.03 & 40.29 & 64.97 & 35.87 & 60.55\\
%Linear after skip & 31.35 & 53.93 & 34.70 & 60.24 & 38.04 & 63.28 & 34.70 & 59.15\\
Static skip & 39.00 & \underline{60.54} & \underline{36.99} & \underline{63.11} & \underline{44.26} & \underline{68.23} & \underline{40.08} & \underline{63.96} \\ %54
%No dropout & \textbf{39.99}& 60.60 & 36.89 & 63.06 & 44.62 & \underline{68.33} & 40.50 & 64.00\\ %15
%No ReLU  & 39.79 & \textbf{61.19} & \underline{37.13} & \textbf{63.31} & \textbf{44.97} & 68.00 & \underline{40.63} & \underline{64.17}\\ %21
\textbf{Proposed Combiner} & \textbf{39.87} & \textbf{60.84} & \textbf{37.67} & \textbf{63.16} & \textbf{44.88} & \textbf{68.59} & \textbf{40.80} & \textbf{64.20}\\\bottomrule
\end{tabular}
\caption{Recall at K on the FashionIQ validation set, with variations on the Combiner architecture. We highlight the best scores in bold and underline the second-best scores.}
\label{tab:ablation_fashion_combiner}
\vspace{-4ex}
\end{table} 

\begin{table}[tb]
\centering
\begin{tabular}{lcc  cccc ccc}
  \toprule
& \multicolumn{4}{c}{\textbf{Recall@K}} & \multicolumn{3}{c}{\textbf{R$_\text{subset}$@K}} \\
\cmidrule(lr){2-5}
\cmidrule(lr){6-8}
\textbf{Model}  & $K=1$ & $K=5$ &  $K=10$ & $K=50$ &  $K=1$ & $K=2$ &  $K=3$ \\
\midrule
Element-wise sum  & 40.97 & 74.70 & 85.51 & 96.94 & 68.81 & 86.96 & 93.90 \\
Convex combination & 41.11 & 75.56 & 85.55 & 97.44 & \textbf{70.46} & 87.08 & \underline{94.33}  \\ %69
W/o convex combination & 36.98 & 72.06 & 82.83 & 96.67 & 65.53 & 84.74 & 93.06 \\ %78
%Linear after skip & 34.41 & 70.05 & 81.75 & 96.29 & 63.62 & 83.57 & 92.97  \\ %129
Static skip & \underline{41.88} & \underline{75.87} & \underline{86.20} & \underline{97.46} & 69.89 & \textbf{87.35} & 94.21 \\ %45
%No dropout & 41.62 & \underline{76.03} & 86.20 & \underline{97.46} & 70.08 & 87.25 & \textbf{94.45} \\ %18
%No ReLU & 41.62 & 75.84 & \underline{86.46} &  97.42 & \textbf{70.35} & \textbf{87.56} & 94.36\\ %12
\textbf{Proposed Combiner} & \textbf{42.05} & \textbf{76.13} & \textbf{86.51} &\textbf{ 97.49} & \underline{70.15} & \underline{87.18} & \textbf{94.40}\\ \bottomrule %39
\end{tabular}
\caption{Recall at K on the CIRR validation set, with variations on the Combiner architecture. We highlight the best scores in bold and underline the second-best scores.}
\label{tab:ablation_cirr_combiner}
\vspace{-4ex}
\end{table} 

We report the results for each variation in \Cref{tab:ablation_fashion_combiner} for the FashionIQ dataset and in \Cref{tab:ablation_cirr_combiner} for the CIRR dataset.
The element-wise sum of the fine-tuned features serves as a solid starting point. As shown in section \ref{sec:fine-tuning}, the task-oriented fine-tuning process is highly effective and results in significant improvements over the out-of-the-box features on both datasets.
The convex combination baseline, which dynamically computes text and image convex coefficients for greater adaptability to the query, achieves a slight improvement over the element-wise sum of the features.
Notably, when we remove the text and image convex combination, we observe a significant drop in performance compared to the proposed Combiner. This emphasizes the importance of the text and image convex combination in achieving good performance.
This result demonstrates that allowing the Combiner network to learn the residual from the element-wise sum (or its generalization, the convex combination) leads to a considerable improvement in performance. This outcome is expected because without the contribution of the image-text convex combination, the effectiveness of the first-stage training, which aims to enhance the additivity properties of the embedding spaces, is compromised.
It is worth noting that setting the convex coefficients statically to 0.5 leads to a slight decrease in performance, which is attributed to the greater adaptability of the dynamically computed coefficients.

Our experiments demonstrate the crucial role of the Combiner architecture in effectively exploiting the full potential of the additive embedding spaces constructed during the first stage of training. By enabling the network to learn the residual from the dynamically computed convex combination, we observe significant performance improvements.

\subsection{Analysis of Two-Stage vs. End-to-End approach}
\begin{table}[tb]
\centering
% \resizebox{1.0\columnwidth}{!}{%
\begin{tabular}{ccc  cc cc cc  cc}
\toprule

&&&\multicolumn{2}{c}{\textbf{Shirt}} & \multicolumn{2}{c}{\textbf{Dress}} & \multicolumn{2}{c}{\textbf{Toptee}} & \multicolumn{2}{c}{\textbf{Average}}\\
\cmidrule(lr){4-5}
  \cmidrule(lr){6-7}
  \cmidrule(lr){8-9}
  \cmidrule(lr){10-11}
\textbf{Approach} & \textbf{IFT} & \textbf{TFT} & $R@10$ & $R@50$ &  $R@10$  & $R@50$ &  $R@10$  & $R@50$ &  $R@10$  & $R@50$  \\
\midrule

\multirow{3}{*}{End-to-end} &\cmark& \xmark & 31.79 & 53.14 & 30.29 & 53.49 & 33.55 & 59.15 & 31.87 & 55.26\\ %10
&\xmark& \cmark & 33.02 & 54.41 & 30.19 & 53.64 & 35.90 & 61.60 & 33.03 & 56.55\\ % 9
&\cmark& \cmark & \underline{37.29} & \underline{59.02} & \underline{34.65} & \underline{60.83} & \underline{41.20} & \underline{65.99} & \underline{37.71} & \underline{61.95}\\ \midrule[.02em] %12

\multirow{3}{*}{Two-stages} &\cmark& \xmark & 34.30 & 55.79 & 32.47 & 55.18 & 38.45 & 62.36 & 35.07 & 57.78\\ %rifare
&\xmark& \cmark & 35.87 & 57.21 & 31.43 & 54.98 & 38.20 & 63.22 & 35.16 & 58.47\\ % 96
&\cmark& \cmark & \textbf{39.87} & \textbf{60.84} & \textbf{37.67} & \textbf{63.16} & \textbf{44.88} & \textbf{68.59} & \textbf{40.80} & \textbf{64.20} \\ \bottomrule %39
\end{tabular}
% }
\caption{Recall at K on the FashionIQ validation set employing either the two-stage or the end-to-end approach. We denote \textbf{IFT} (image encoder fine-tuning) and \textbf{TFT} (text encoder fine-tuning) to represent whether the image encoder or the text encoder is fine-tuned in the first stage. We highlight the best scores in bold and underline the second-best scores.}
\label{tab:fashioniq-stages}
\vspace{-3ex}
\end{table}

\begin{table}[tb]
\centering
% \resizebox{1.00\columnwidth}{!}{%
\begin{tabular}{ccc  cccc ccc}
  \toprule

&&& \multicolumn{4}{c}{\textbf{Recall@K}} & \multicolumn{3}{c}{\textbf{R$_\text{subset}$@K}} \\
\cmidrule(lr){4-7}
\cmidrule(lr){8-10}
\textbf{Approach} & \textbf{IFT} & \textbf{TFT}  & $K=1$ & $K=5$ &  $K=10$ & $K=50$ &  $K=1$ & $K=2$ &  $K=3$ \\
\midrule
\multirow{3}{*}{End-to-end}&\cmark& \xmark & 33.19 & 67.01 & 79.83 & 95.52 & 58.90 & 79.33 & 90.28 \\  %19
&\xmark& \cmark & 33.58 & 67.59 & 79.57 & 95.28 & 67.37 & 85.58 & 93.44 \\ %10~
&\cmark& \cmark & \underline{40.03} & \underline{74.09} & \underline{85.14} & \underline{97.12} & 68.14 & 86.06 & 93.64\\ \midrule[.02em] %20

\multirow{3}{*}{Two-stages}&\cmark& \xmark & 34.01 & 69.07 & 81.77 & 95.72 & 62.78 & 81.80 & 91.41 \\ %45
&\xmark& \cmark & 36.86 & 71.32 & 82.32 & 96.24 & \underline{68.28} & \underline{86.51} & \underline{94.14} \\ %54
&\cmark& \cmark & \textbf{42.05} & \textbf{76.13} & \textbf{86.51} & \textbf{97.49} & \textbf{70.15} & \textbf{87.18} & \textbf{94.40}\\ \bottomrule % 33
\end{tabular}
% }
\caption{Recall at K on the CIRR validation set employing either the two-stage or the end-to-end approach. We denote \textbf{IFT} (image encoder fine-tuning) and \textbf{TFT} (text encoder fine-tuning) to represent whether the image encoder or the text encoder is fine-tuned in the first stage. We highlight the best scores in bold and underline the second-best scores.}
\label{tab:cirr-stages}
\vspace{-4ex}
\end{table}
In order to explain why a two-stage training method, where the CLIP encoder and Combiner are trained separately, in contrast to an end-to-end approach, we perform an experiment where we compare the two settings on both CIRR and FashionIQ datasets. First, we train end-to-end by fine-tuning CLIP encoders while training the Combiner network simultaneously. Then, we followed the proposed two-stage approach. In both settings, we also enable fine-tuning of the textual or image encoders separately and jointly. In all the experiments in this section, we use the RN-50 CLIP model.

We present the results in \Cref{tab:fashioniq-stages} for the FashionIQ dataset and in \Cref{tab:cirr-stages} for the CIRR dataset. Remarkably, the two-stage approach consistently outperforms the end-to-end one on both datasets. These superior results remain consistent even when varying the fine-tuning modality.

The results validate the effectiveness of constructing an embedding space with robust additivity properties before combining the features using a non-linear function. We hypothesize that when training the Combiner network simultaneously with the CLIP encoders, the entire system struggles to effectively learn the additive embedding spaces and the non-linear combining function in a cohesive manner. As a result, this limitation negatively impacts the overall performance, leading to suboptimal outcomes.

\subsection{Preprocess upshot}

In this section, we show how the proposed preprocess pipeline, described in \Cref{sec:preprocess}, contributes to further improving performance. We compare the proposed preprocess with two other methods: the standard CLIP preprocess pipeline, primarily consisting of resize and center crop operations, and the Square preprocess, which involves applying a square zero-pad to the image before resizing and center cropping.
The comparison among the different preprocess techniques is presented in \Cref{tab:fashioniq-preprocess} for the FashionIQ dataset and \Cref{tab:cirr-preprocess} for the CIRR dataset.

On the FashionIQ dataset, the improvement obtained using the proposed preprocess pipeline over the standard one is substantial in the \emph{Dress} category and noticeable in the \emph{Toptee} category.
Conversely, the square pad preprocess technique achieves comparable performance to the proposed one in the \emph{Dress} and \emph{Toptee} categories while suffering a performance deficit in the \emph{Shirt} category.
Overall, we observe a correlation between the difference in performance among the methods and the number of images with a high aspect ratio, as depicted in \Cref{fig:hist_plot}. 
In other words, when dealing with images with a high aspect ratio, it is preferable to pad them to avoid losing crucial portions of the image during the center crop operation. On the other hand, when images have a low aspect ratio, it is more effective not to reduce the usable portion of the image with padding. The proposed preprocess pipeline achieves the best performance by effectively adapting to the aspect ratio of each image.
On the CIRR dataset, we observe that the proposed preprocess significantly improves performance compared to the standard CLIP and the square preprocess. The performance gain is particularly significant in low-rank recall measures, where the importance of every lost detail is crucial.
\begin{table}[tb]
\centering
% \resizebox{1.0\columnwidth}{!}{%
\begin{tabular}{ccc  cc cc cc  cc}
\toprule
&&\multicolumn{2}{c}{\textbf{Shirt}} & \multicolumn{2}{c}{\textbf{Dress}} & \multicolumn{2}{c}{\textbf{Toptee}} & \multicolumn{2}{c}{\textbf{Average}}\\
\cmidrule(lr){3-4}
\cmidrule(lr){5-6}
  \cmidrule(lr){7-8}
  \cmidrule(lr){9-10}
\textbf{CF} & \textbf{Preprocess} & $R@10$ & $R@50$ &  $R@10$  & $R@50$ &  $R@10$  & $R@50$ &  $R@10$  & $R@50$  \\
\midrule
\multirow{3}{*}{Sum} & \textbf{Standard} & 37.64 & 59.76 & 33.42 & 59.84 & 40.90 & 66.80 & 37.32 & 62.13\\ %54
&\textbf{Square} & 37.09 & 58.52 & 35.94 & 62.03 & 42.53 & 66.29 & 38.52 & 62.28\\ %12
&\textbf{Proposed}  & 38.67 & 59.42 & 35.99 & 62.22 & 43.35 & 67.52 & 39.34 & 63.05 \\ \midrule[.02em]
\multirow{3}{*}{Combiner} & \textbf{Standard} & \underline{39.40} & \textbf{61.33} & 35.25 & 60.44 & 43.95 & 67.72 & 39.53 & 63.16\\ %54
& \textbf{Square} & 38.71 & 60.21 & \textbf{37.97} & \underline{62.86} & \underline{44.12} & \underline{68.03} & \underline{40.26} & \underline{63.70}\\ %48?
& \textbf{Proposed} & \textbf{39.87} & \underline{60.84} & \underline{37.67} & \textbf{63.16} & \textbf{44.88} & \textbf{68.59} & \textbf{40.80} & \textbf{64.20} \\ \bottomrule
\end{tabular}
% }

\caption{Recall at K on FashionIQ validation set varying the combining function and the preprocessing pipeline used. \textbf{CF} (combining function) indicates the function used to combine the query features. We highlight the best scores in bold and underline the second-best scores.}
\label{tab:fashioniq-preprocess}
\vspace{-3ex}
\end{table}

\begin{table}[tb]
\centering
% \resizebox{1.00\columnwidth}{!}{%
\begin{tabular}{ccc  cccc ccc}
  \toprule
&& \multicolumn{4}{c}{\textbf{Recall@K}} & \multicolumn{3}{c}{\textbf{R$_\text{subset}$@K}} \\
\cmidrule(lr){3-6}
\cmidrule(lr){7-9}
\textbf{CF} & \textbf{Preprocess}  & $K=1$ & $K=5$ &  $K=10$ & $K=50$ &  $K=1$ & $K=2$ &  $K=3$ \\
\midrule
\multirow{3}{*}{Sum} & \textbf{Standard} & 39.51 & 74.00 & 84.72 & \underline{97.20} & 68.36 & 86.15 & 94.26 \\ %18
&\textbf{Square} & 41.26 & 74.34 & 85.00 & 96.84 & 69.15 & 85.89 & 93.90 \\ %?
&\textbf{Proposed}  & 40.97 & 74.70 & \underline{85.51} & 96.94 & 68.81 & \underline{86.96} & 93.90 \\ \midrule[.02em]
\multirow{3}{*}{Combiner} & \textbf{Standard} &  40.08 & 74.15 & 84.67 & \underline{97.20} & 69.53 & 86.27 & \textbf{94.45} \\ %18
&\textbf{Square} & \underline{41.95} & \underline{74.96} & 85.24 & 96.58 & \textbf{70.65} & 86.67 & 94.24 \\ %21
& \textbf{Proposed} & \textbf{42.05} & \textbf{76.13} & \textbf{86.51} & \textbf{97.49} & \underline{70.15} & \textbf{87.18} & \underline{94.40} \\ \bottomrule
\end{tabular}
% }
\caption{Recall at K on CIRR validation set varying the combining function and the preprocessing pipeline used. \textbf{CF} (combining function) indicates the function used to combine the query features. We highlight the best scores in bold and underline the second-best scores.}
\label{tab:cirr-preprocess}
\vspace{-4ex}
\end{table}

\subsection{Comparison with SotA}
We compare the proposed method with state-of-the-art approaches on two standard and challenging datasets.
To ensure a fair comparison, we follow the standard experimental settings of the two datasets \cite{wu2021fashion, liu2021image}. 
Unless specifically mentioned, we report the metrics for each method as documented in the official papers, and we refer to those papers for more comprehensive details about the individual approaches.

\begin{table}[t]
\centering
\resizebox{1.0\columnwidth}{!}{%
\begin{tabular}{lcccc  cc cc cc  cc}
\toprule
&\multicolumn{2}{c}{\textbf{Encoder}}&\multicolumn{2}{c}{\textbf{Shirt}} & \multicolumn{2}{c}{\textbf{Dress}} & \multicolumn{2}{c}{\textbf{Toptee}} & \multicolumn{2}{c}{\textbf{Average}}\\
  \cmidrule(lr){2-3}
\cmidrule(lr){4-5}
  \cmidrule(lr){6-7}
  \cmidrule(lr){8-9}
  \cmidrule(lr){10-11}
\textbf{Method}& Visual & Textual & $R@10$ & $R@50$ &  $R@10$  & $R@50$ &  $R@10$  & $R@50$ &  $R@10$  & $R@50$  \\
\midrule
%JVSM~\cite{Chen2020jvsm} & MobileNet~\cite{howard2017mobilenets} & LSTM~\cite{hochreiter1997long} & 12.0 & 27.1 & 10.7 & 25.9 & 13.0 & 26.9 & 11.9 & 26.6  \\
TRACE~\cite{jandial2020trace} & RN-50 & BERT~\cite{devlin2019bert}& 20.80 & 40.80 & 22.70 & 44.91 & 24.22 & 49.80 & 22.57 & 46.19  \\ 
VAL~\cite{Chen_2020_CVPR} & RN-50 & LSTM(GloVe)~\cite{hochreiter1997long} & 22.38 & 44.15 & 22.53 & 44.00 & 27.53 & 51.68 & 24.15 & 46.61  \\ 
%MAAF \cite{dodds2020modality} & 21.3 & 44.2 & 23.8 & 48.6 & 27.9 & 53.6 & 24.3 &  48.8\\ 
CurlingNet~\cite{yu2020curlingnet} & RN-152 & biGRU~\cite{chung2014empirical}& 21.45 & 44.56 & 26.15 & 53.24 & 30.12 & 55.23 & 25.90 & 51.01  \\ 
RTIC-GCN~\cite{shin2021rtic} & RN-50 & LSTM(GloVe) & 23.79 & 47.25 & 29.15 & 54.04 & 31.61 & 57.98 & 28.18 & 53.09  \\ 
CoSMo~\cite{Lee_2021_CVPR} & RN-50 & LSTM & 24.90 & 49.18 & 25.64 & 50.30 & 29.21 & 57.46 & 26.58 & 52.31 \\ 
DCNet~\cite{Kim_Yu_Kim_Kim_2021} & RN-50 & Conv1D(GloVe) & 23.95 & 47.30 & 28.95 & 56.07 & 30.44 & 58.29 & 27.78 & 53.89 \\ 
CLVC-Net~\cite{clvc-net} & RN-50 & LSTM & 28.75 & 54.76 & 29.85 & 56.47 & 33.50 & 64.00 & 30.70 & 58.41  \\ 
\midrule[.02em]
CIRPLANT~\cite{liu2021image} & RN-152 & OSCAR~\cite{li2020oscar} & 17.53 & 38.81 & 17.45 & 40.41 & 21.64 & 45.38 & 18.87 & 41.53 \\
MAAF~\cite{dodds2020modality} & RN-50 & BERT & 18.55 & 37.63 & 18.59 & 39.66 & 23.05 & 45.95 & 20.06 & 41.08\\ 
SAC~\cite{Jandial_2022_WACV} & RN-50 & BERT & 28.02 & 51.86 & 26.52 & 51.01 & 32.70 & 61.23 & 29.08 & 54.70 \\ 
FashionViL~\cite{han2022fashionvil} & RN-50 & BERT & 25.17 & 50.39 & 33.47 & 59.94 & 34.98 & 60.79 & 31.20 & 57.04 \\
\rowcolor{LightCyan}
\textbf{Ours} & RN-50 & Transformer & \underline{39.87} & \underline{60.84} & \underline{37.67} & \underline{63.16} & \underline{44.88} & \underline{68.59} & \underline{40.80} & \underline{64.20} \\ 
\rowcolor{LightCyan}
\textbf{Ours} & RN-50x4 & Transformer & \textbf{44.41} & \textbf{65.26} & \textbf{39.46} & \textbf{64.55} & \textbf{47.48} & \textbf{70.98} & \textbf{43.78} & \textbf{66.93}\\ \bottomrule %150
\end{tabular}
}
\caption{Comparison between our method and current state-of-the-art models on the Fashion-IQ validation set. We highlight the best scores in bold and underline the second-best scores. The upper section of the table presents methods that are not directly comparable to our proposed approach, as they either do not utilize a pre-trained textual encoder or do not update its weights. "RN" stands for ResNet.}
\label{tab:fashioniq-comparison}
\vspace{-4ex}
\end{table}

\Cref{tab:fashioniq-comparison} reports the comparison between the proposed method and other state-of-the-art approaches. 
We divide the table into two sections: the upper section includes methods that are not directly comparable to our approach. These approaches either do not utilize a pre-trained textual encoder~\cite{Chen_2020_CVPR, yu2020curlingnet, shin2021rtic, Lee_2021_CVPR, Kim_Yu_Kim_Kim_2021,clvc-net} or, in the case of TRACE~\cite{jandial2020trace}, they use BERT~\cite{devlin2019bert} as a pre-trained textual encoder but do not update its weights. 
It is important to note that even when a competitor \cite{Chen_2020_CVPR, shin2021rtic, Kim_Yu_Kim_Kim_2021} utilizes the GloVe word embedding~\cite{pennington2014glove}, we do not consider their textual encoder as pre-trained.
All the methods in this section rely on a ResNet model pre-trained on the ImageNet dataset~\cite{russakovsky2015imagenet} and fine-tuned during training. We include the results of these methods to provide a more comprehensive discussion.
The lower section of \Cref{tab:fashioniq-comparison} reports methods that are directly comparable to ours: they rely on both pre-trained visual and language models updating all the weights of both backbones during training.
CIRRPLANT~\cite{liu2021image} relies on the OSCAR pretrained model as a textual backbone, while ~\cite{dodds2020modality, Jandial_2022_WACV, han2022fashionvil} rely on the pre-trained BERT model. It is worth mentioning that FashionViL is a fashion-oriented approach that carries out a large-scale pre-training for learning V+L representation in the fashion domain. For this reason, it is not surprising that it exhibits strong performances in a fashion dataset such as FashionIQ.
When considering the RN50-based method, the proposed approach outperforms the competitors by improving up to 9\% in average R@10 and 7\% in average R@50 compared to the best-performing competitor, FashionViL, when using the same visual backbone architecture. Our method demonstrates the highest recall across all categories, with a particularly significant margin observed in the Shirt category.
When considering the larger RN-50x4-based model, we observe an improvement ranging from 2\% to 4\% in all categories compared to the smaller backbone. This result demonstrates that our approach scales well when using larger and heavier VL models.

\begin{table}[t]
\centering
\begin{tabular}{lcccc  cccc ccc}
  \toprule
& \multicolumn{2}{c}{Encoder} &\multicolumn{4}{c}{\textbf{Recall@K}} & \multicolumn{3}{c}{\textbf{R$_\text{subset}$@K}} \\
\cmidrule(lr){2-3}
\cmidrule(lr){4-7}
\cmidrule(lr){8-10}
\textbf{Method} & Visual & Textual & $K=1$ & $K=5$ &  $K=10$ & $K=50$ &  $K=1$ & $K=2$ &  $K=3$ \\
\midrule
TIRG$\textsuperscript{\dag}$ \cite{vo2019composing} & RN-18 & LSTM & 14.61 & 48.37 & 64.08 & 90.03 & 22.67 & 44.97 & 65.14   \\
TIRG$+$LastConv$\textsuperscript{\dag}$ \cite{vo2019composing} & RN-18 & LSTM & 11.04 & 35.68 & 51.27 & 83.29 & 23.82 & 45.65 & 64.55   \\ 
MAAF$\textsuperscript{\dag}$ \cite{dodds2020modality} & RN-50 & LSTM & 10.31 & 33.03 & 48.30 & 80.06 & 21.05 & 41.81 & 61.60 \\ 
MAAF$-$IT$\textsuperscript{\dag}$ \cite{dodds2020modality} & RN-50 & LSTM & 9.90 & 32.86 & 48.83 & 80.27 & 21.17 & 42.04 & 60.91 \\ 
MAAF$-$RP$\textsuperscript{\dag}$ \cite{dodds2020modality} & RN-50 & LSTM & 10.22 & 33.32 & 48.68 & 81.84 & 21.41 & 42.17 & 61.60 \\ 
ARTEMIS \cite{delmas2021artemis} & RN-152 & biGRU & 16.96 & 46.10 & 61.31 & 87.73 & 39.99 & 62.20 & 75.67 \\ \midrule[.02em]
% CIRPLANT $\textsuperscript{\dag}$ \cite{liu2021image} & RN-152 & Transformer & 15.18 & 43.36 & 60.48 & 87.64 & 33.81 & 56.99 & 75.40 \\ \midrule[.02em]
MAAF$\textsuperscript{\dag}$ \cite{dodds2020modality} & RN-50 & BERT & 10.12 & 33.10 & 48.01 & 80.57 & 22.04 & 42.41 & 62.14 \\
CIRPLANT$\textsuperscript{\dag}$ \cite{liu2021image} & RN-152 & OSCAR & 19.55 & 52.55 & 68.39 & 92.38 & 39.20 & 63.03 & 79.49 \\
\rowcolor{LightCyan}
\textbf{Ours} & RN-50 & Transformer & \underline{40.91} & \underline{74.53} & \underline{84.77} & \underline{97.35} & \underline{70.22} & \underline{87.80} & \underline{94.46} \\ 
\rowcolor{LightCyan}
\textbf{Ours} & RN-50x4 & Transformer & \textbf{44.82} & \textbf{77.04} & \textbf{86.65} & \textbf{97.90} & \textbf{73.16} & \textbf{88.84} & \textbf{95.59}\\ \bottomrule
\end{tabular}
\caption{Comparison between our method and current state-of-the-art models on the CIRR test set. We highlight the best scores in bold and underline the second-best scores.. $\textsuperscript{\dag}$ denotes results cited from \cite{liu2021image}. The upper section of the table presents methods that are not directly comparable to our proposed approach, as they either do not utilize a pre-trained textual encoder or do not update its weights. In the lower section, we report methods that are directly comparable to our approach. "RN" stands for ResNet.}
\label{tab:cirr-comparison}
\vspace{-4ex}
\end{table}

In \Cref{tab:cirr-comparison}, we report a comparison between the proposed method and other state-of-the-art approaches. As for FashionIQ, the upper section of the table reports methods that are not directly comparable with the proposed one: they do not utilize a pre-trained textual encoder. As the visual backbone, they employ a ResNet-based model, which is pre-trained on ImageNet and fine-tuned during training.
The lower section of the table includes directly comparable methods, such as MAAF, which utilizes BERT as a text encoder, and CIRPLANT, which relies on the pre-trained Vision-Language model OSCAR. The results presented in \Cref{tab:cirr-comparison} are obtained through the official evaluation server. Our approach consistently outperforms the competitors by a significant margin, particularly in low-rank recall measures, where we notice an improvement of approximately 20\% in R@1 when using the RN50 visual backbone. When considering the larger RN-50x4 model, we observe improvements ranging from 3\% in low-rank recall metrics to 1\% as the recall rank increases.

% \Cref{tab:cirr-comparison} shows the quantitative results on the CIRR test set obtained through the official evaluation server. Also in this dataset, our approach consistently outperforms current methods by a large margin, especially in low-rank recall measures where we achieve an improvement up to $\sim25\%$ in R@1. Also, the results of the retrieval within the subset of the queries are very good, with an improvement up to $\sim33\%$ in R$_\text{Subset}$@1.
% The gap between the two backbones is similar to that observed in FashionIQ: it is in the range of about 3\%, observed in the low-rank recall metrics, to 1\% when the rank of the recall metrics increases. 

\subsection{Feature distribution study}\label{sec:feat_study}
\begin{figure}[t]
    \centering
    \begin{subfigure}{0.48\linewidth}
    \includegraphics[width=\linewidth]{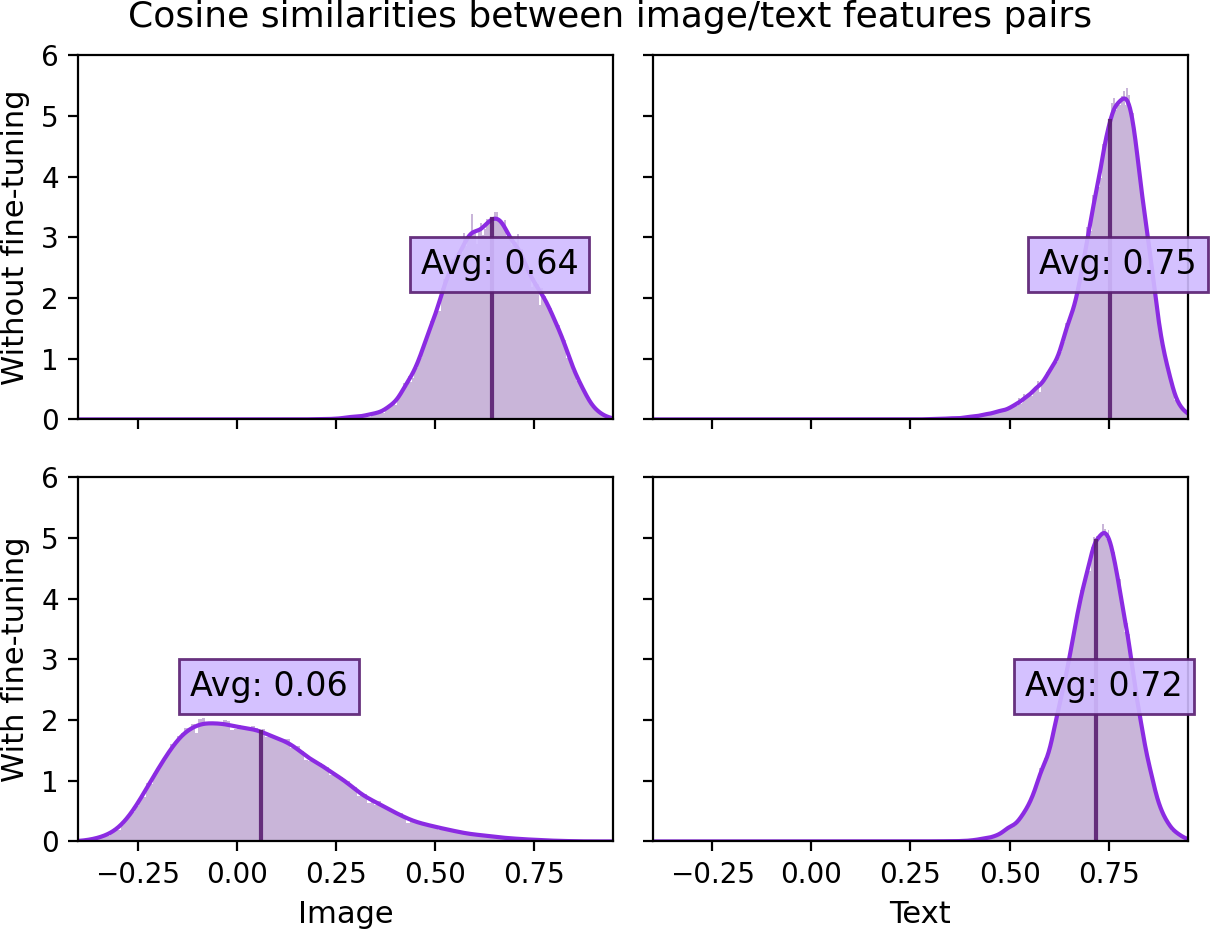}
    \caption{FashionIQ dataset}
    \label{fig:pairwise-fiq}
    \end{subfigure}\hfill
 \begin{subfigure}{0.48\linewidth}
    \includegraphics[width=\linewidth]{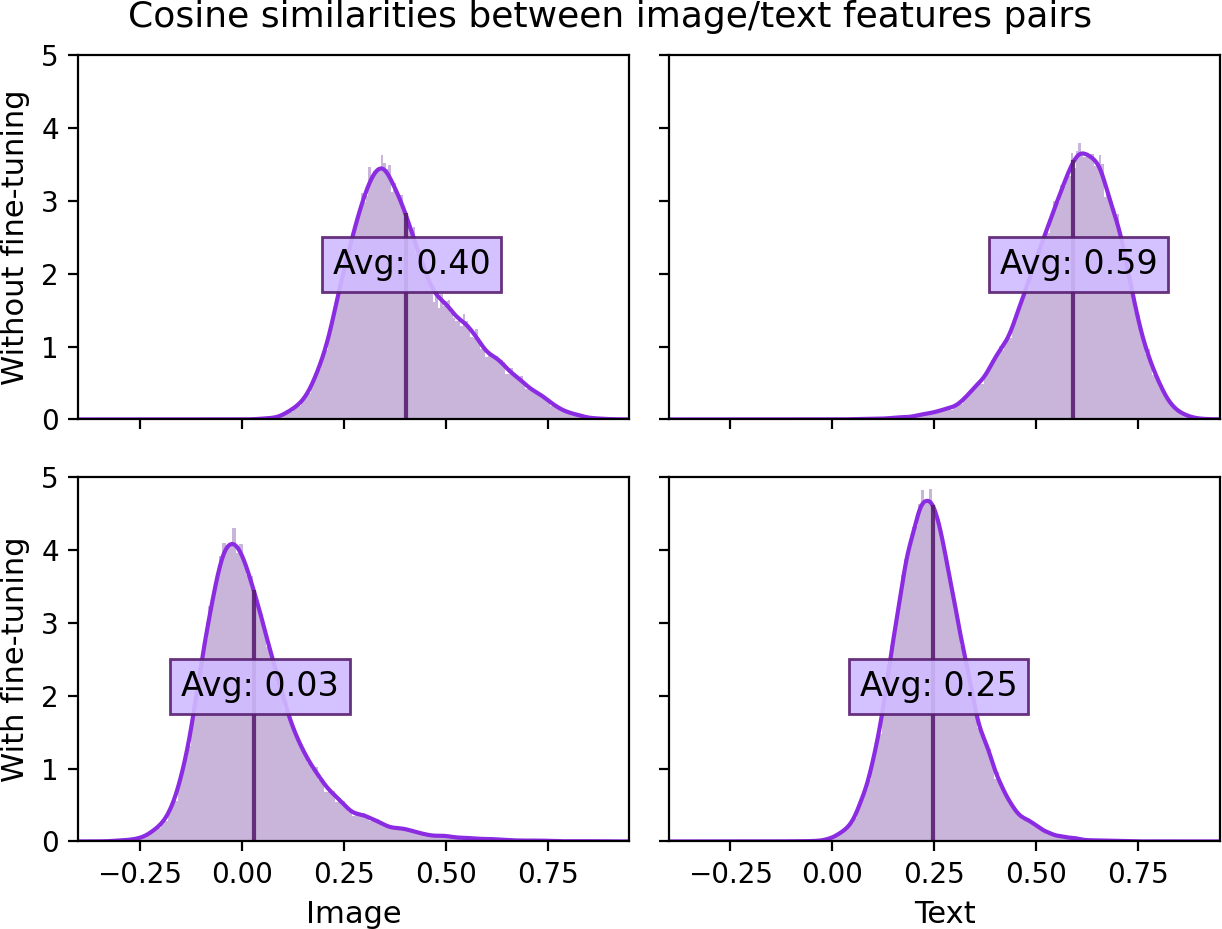}
    \caption{CIRR dataset}
    \label{fig:pairwise-cirr}
    \end{subfigure}
    \vspace{-1.8ex}
    \caption{Histograms of cosine similarities between image/text feature pairs. The x-axis represents the cosine similarities. The y-axis represents the (normalized) number of pairs.
    In the top-line plots, we have used the out-of-the-box CLIP model. In the bottom line, we have used the model fine-tuned during the first stage of training. In the left-side plots, we compare the image features. In the right ones, we compare the text features. The histograms are normalized such that the area under each curve integrates to 1.}
    \label{fig:pairwise-similarities}
    \vspace{-2.5ex}
\end{figure}
The experiments in this section aim to provide intuition on how the feature distribution in the embedding spaces affects the retrieval performances. All the experiments were carried out on the validation sets using the RN-50 model.
We are going to present two different sets of experiments that have slightly different purposes. The first set aims to investigate how the image and text features are distributed in the embedding spaces, while the second one explores how the distribution of the features affects retrieval performance.

To investigate how the features distribute in the embedding spaces, we followed \cite{liang2022mind} and calculate pairwise similarities among them. If the features occupy the embedding space uniformly, their similarities will be lower. Throughout all experiments, due to the quadratic growth of possible pairs, we compute the similarities between 50K randomly sampled pairs.
\Cref{fig:pairwise-similarities} shows the histograms of the features pairwise similarities on both FashionIQ and CIRR datasets.
First of all, we can notice that due to the broader domain of CIRR, on such a dataset, both the image and text features similarities are higher when compared to FashionIQ.
On both datasets, fine-tuning the image encoder leads to a drastic reduction in the average similarity of the visual features and, thus, to much more efficient use of the embedding space during retrieval.
This fact confirms our hypothesis (\Cref{sec:fine-tuning}) that fine-tuning the image encoder adapts the image manifold to the data domain.
The fine-tuning of the text encoder leads to a lower reduction in the average pairwise similarity of textual features (almost negligible in FashionIQ) than that observed in visual ones. 
We suppose that efficient use of the image embedding space is far more critical than efficient use of the textual space since the retrieval is carried out in the image space.
In all the experiments, we observe that the fine-tuning of CLIP encoders contributes to reducing the \textit{cone effect}: ``the effective embedding space is restricted to a narrow cone for trained models and models with random weights" \cite{liang2022mind}.

The previous experiments demonstrate how the two-stage approach proposed in this study affects the textual and visual CLIP embedding spaces. However, these experiments do not clarify why this increased utilization of embedding space can improve the retrieval process. We conducted additional experiments to investigate the impact of this embedding space reshaping on the image retrieval task.
We compute and compare the cosine similarities (the distance function used in the retrieval) between the combined and the index image features. Specifically, we perform two distinct computations: in the first one, we compute the similarity between the combined features and the target image features belonging to the same query triplet. In the second one, we compute the similarity between the combined features and random image features that differ from the target ones. Given a query, we will refer to the images that differ from the target as \textit{non-target images}. We compare each combined feature with ten non-target image features to reduce the variance.

\begin{figure}[t]
    \centering
    \begin{subfigure}{0.48\linewidth}
    \includegraphics[width=\linewidth]{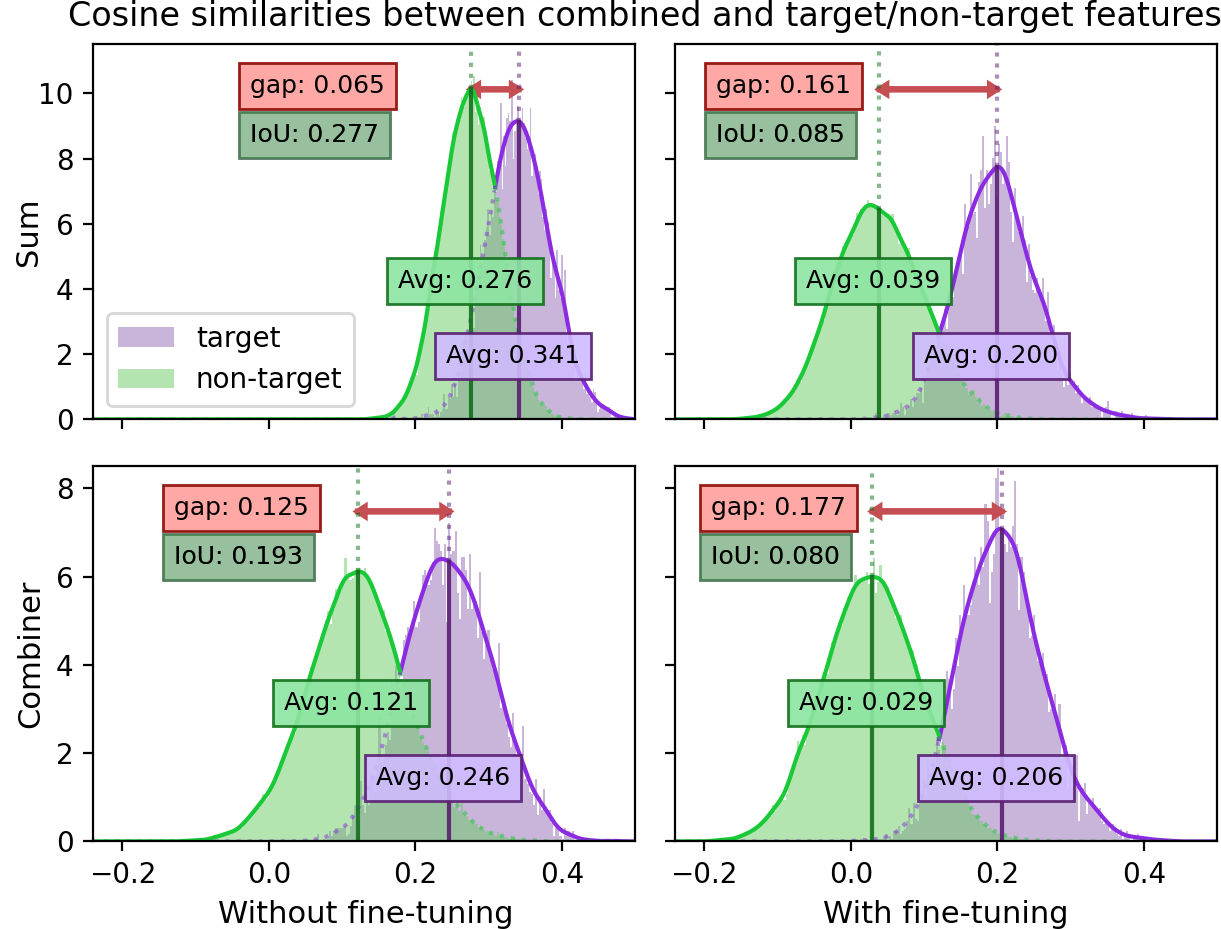}
    \caption{FashionIQ dataset}
    \label{fig:predicted-fiq}
    \end{subfigure}\hfill
 \begin{subfigure}{0.48\linewidth}
    \includegraphics[width=\linewidth]{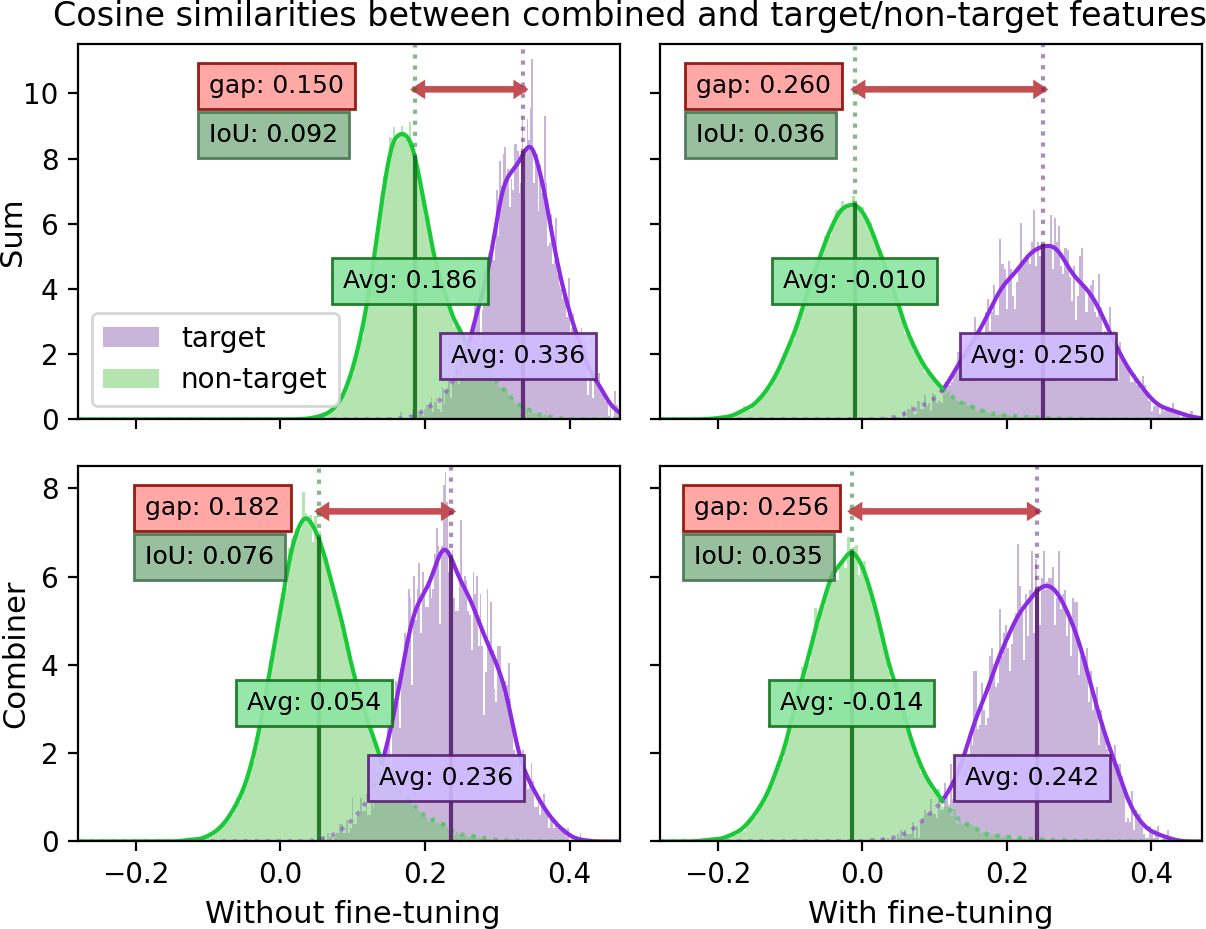}
    \caption{CIRR dataset}
    \label{fig:predicted-cirr}
    \end{subfigure}
    \vspace{-1.8ex}
    \caption{Histograms of the cosine similarities between combined and target/non-target feature pairs.
    The x-axis represents the cosine similarities. The y-axis represents the (normalized) number of pairs. In \mbox{green{\LARGE\color{plotgreen}\textbullet}}: cosine similarities between combined and non-target index features. In \mbox{violet{\LARGE\color{plotviolet}\textbullet}}: cosine similarities between combined and target features. In \mbox{red{\LARGE\color{plotred}\textbullet}}: similarity gap between combined-target and combined-non target features. In \mbox{dark green{\LARGE\color{plotdarkgreen}\textbullet}}: intersection over union area between the two histograms.
    In the top-line plots, we have used the simple sum as a combining function. In the bottom line ones, we have used the Combiner network. In the left-side plots, we used the out-of-the-box CLIP model. In the right ones, we used the model fine-tuned during the first stage of the training. 
    The histograms are normalized such that the area under each curve integrates to 1.}
    \label{fig:predicted-similarities}
    \vspace{-2ex}
\end{figure}

\Cref{fig:predicted-similarities} emphasizes the similarity gaps between the combined and target/non-target features. On both FashionIQ and CIRR datasets, we notice that the element-wise sum of out-of-the-box CLIP features achieves the highest average combined-target features similarity. During both the fine-tuning and the Combiner network training stages, the contrastive training increases the cosine distances between the combined and non-target features instead of increasing their similarity to the target features.
By observing both \cref{fig:predicted-fiq} and \cref{fig:predicted-cirr} and the corresponding retrieval results in \Cref{tab:fashioniq-finetune} and \Cref{tab:cirr-finetune}, we argue that, in these two datasets, the retrieval performances are highly correlated with the similarity gap between combined-target and combined-non target features (displayed as the red arrows in \cref{fig:predicted-similarities}) and with the size of intersection area between the histograms (the smaller the intersection area, the smaller the retrieval errors will be). On the contrary, the absolute value of the combined-target similarity does not seem to be of great importance.

The two sets of experiments highlight different but strongly related aspects. The first set shows that fine-tuning both CLIP encoders leads to more efficient use of the embedding spaces. In the second set, we prove that the increased occupation of the image space helps to ``move away" the combined features from the non-target features.

\subsection{Qualitative results}
To obtain a clearer understanding of which parts of the images the system considers most important during retrieval, we conducted qualitative experiments using the GradCAM technique \cite{Selvaraju2019gradCAM}.
Instead of computing gradients versus an output class, we compute gradients with respect to the combined features, which summarize both the visual and textual content of the image and caption, using the GradCAM technique. This approach makes each heat map generated by GradCAM dependent on the reference image and its relative caption, simulating the retrieval process. We use the last convolutional layer of CLIP's image encoder as the saliency layer.

\begin{figure*}[t]
    \centering
    \includegraphics[width=0.95\textwidth]{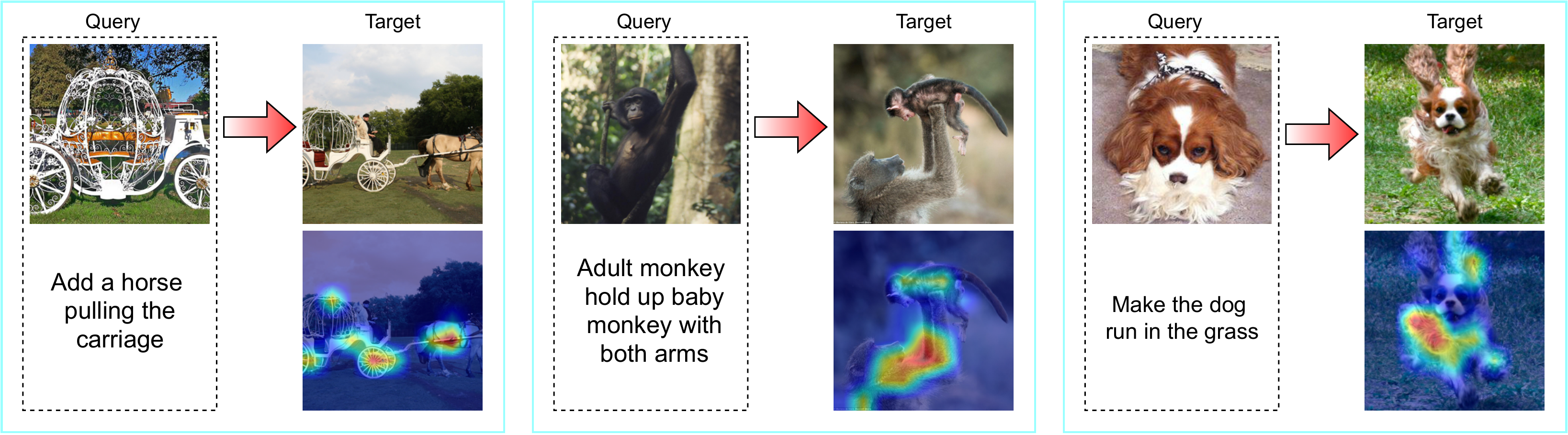}
    \vspace{-1ex}

    \caption{Examples of GradCAM visualization on CIRR dataset computing the gradients with respect to the Combiner output.}
    \label{fig:cirr_gradcam}
    \vspace{-2ex}
\end{figure*}

\begin{figure*}[t]
    \centering
    \includegraphics[width=0.95\textwidth]{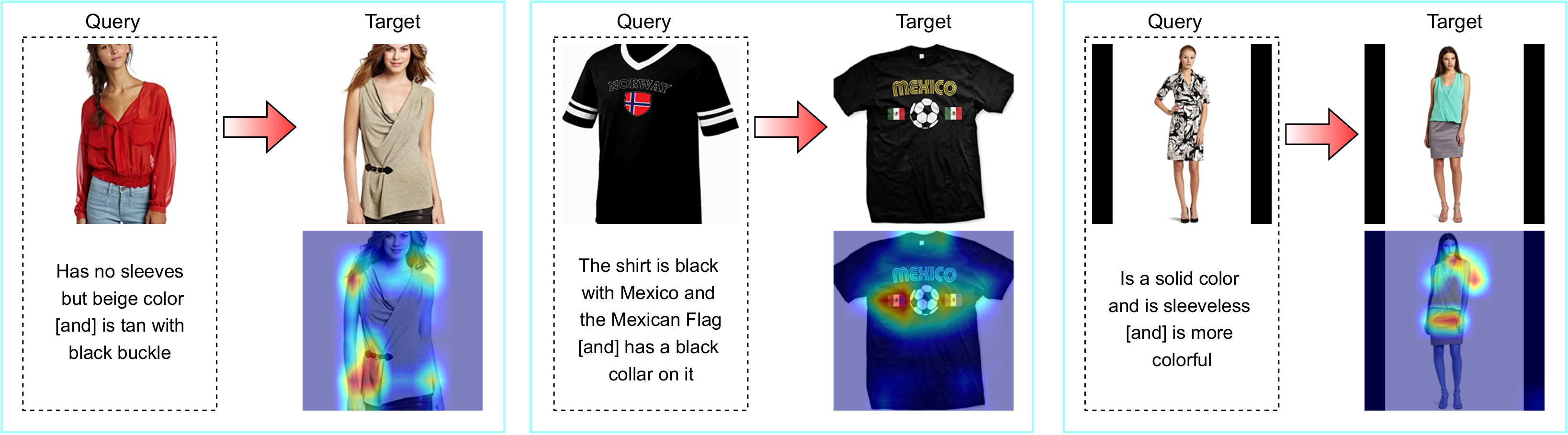}
        \vspace{-1ex}
    \caption{Examples of GradCAM visualization on FashionIQ computing the gradients with respect to the Combiner output.}
    \label{fig:fashioniq_gradcam}
    \vspace{-3ex}

\end{figure*}

In \cref{fig:cirr_gradcam} and in \cref{fig:fashioniq_gradcam} are displayed some examples of the above-described visualization technique.
The system is capable of attending to a wide range of concepts, such as style and color changes for the fashion dataset and behavior modification for the CIRR dataset, as we can notice from the experiments with the GradCAM technique. For instance, in \cref{fig:cirr_gradcam}, the system attends to the carriage and horse in the first example, the pose of the holding monkey and the baby monkey in the second example, and the pose of the dog in the third example. In \cref{fig:fashioniq_gradcam}, the system attends to the arms and shoulders of the person when the conditioning text referred to the sleeves of the dress and to the logo of the shirt when it was requested to change the Norwegian flag into a Mexican one. 

\begin{figure*}[!htb]
    \centering
    \includegraphics[width=\textwidth]{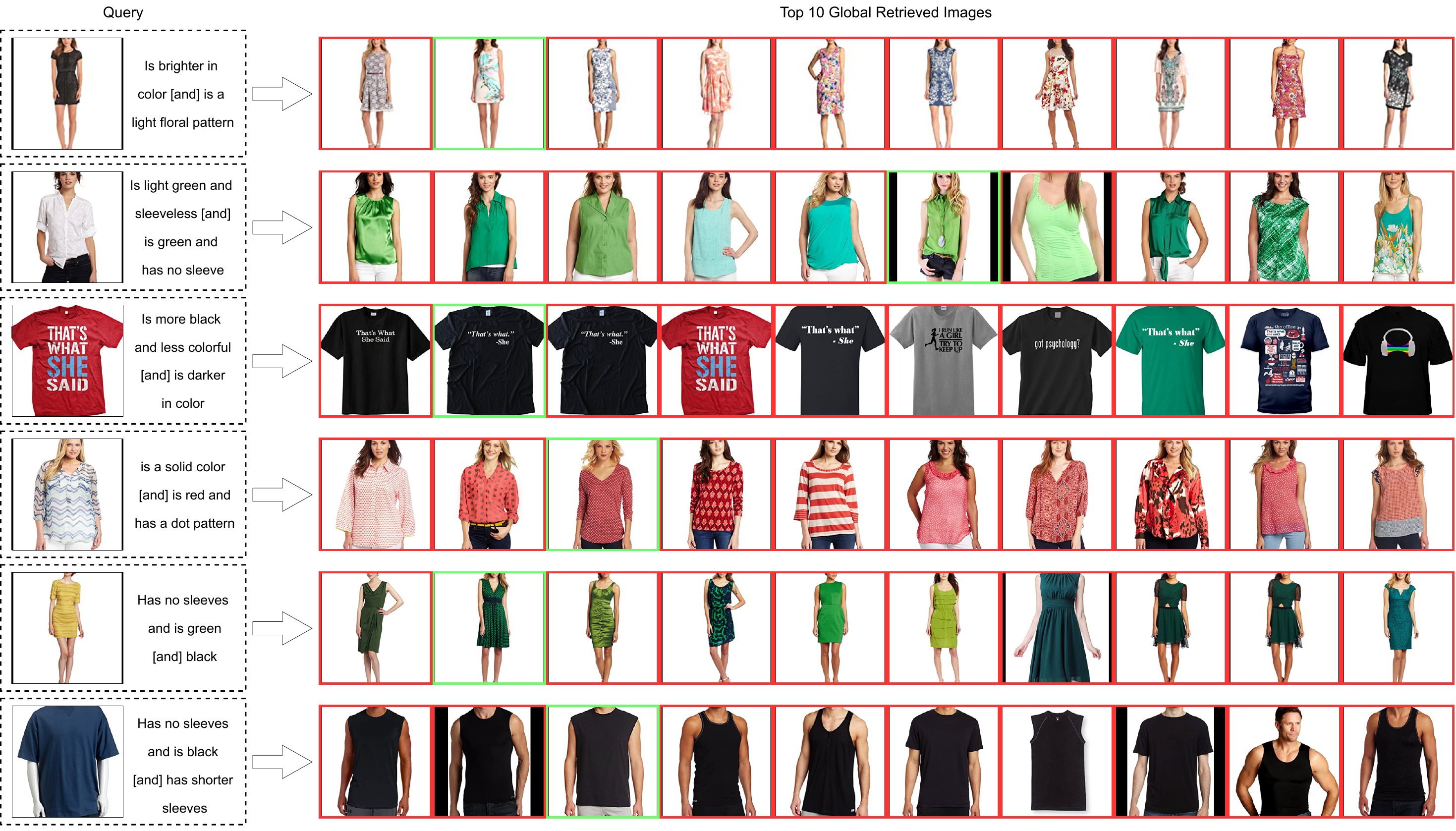}
    \caption{Qualitative results for the FashionIQ dataset. We highlight with a green border when the retrieved image is labeled as ground truth for the given query.}
    \label{fig:fashion_retrieved}
    \vspace{-2ex}
\end{figure*}

\begin{figure*}[!htb]
    \centering
    \includegraphics[width=\textwidth]{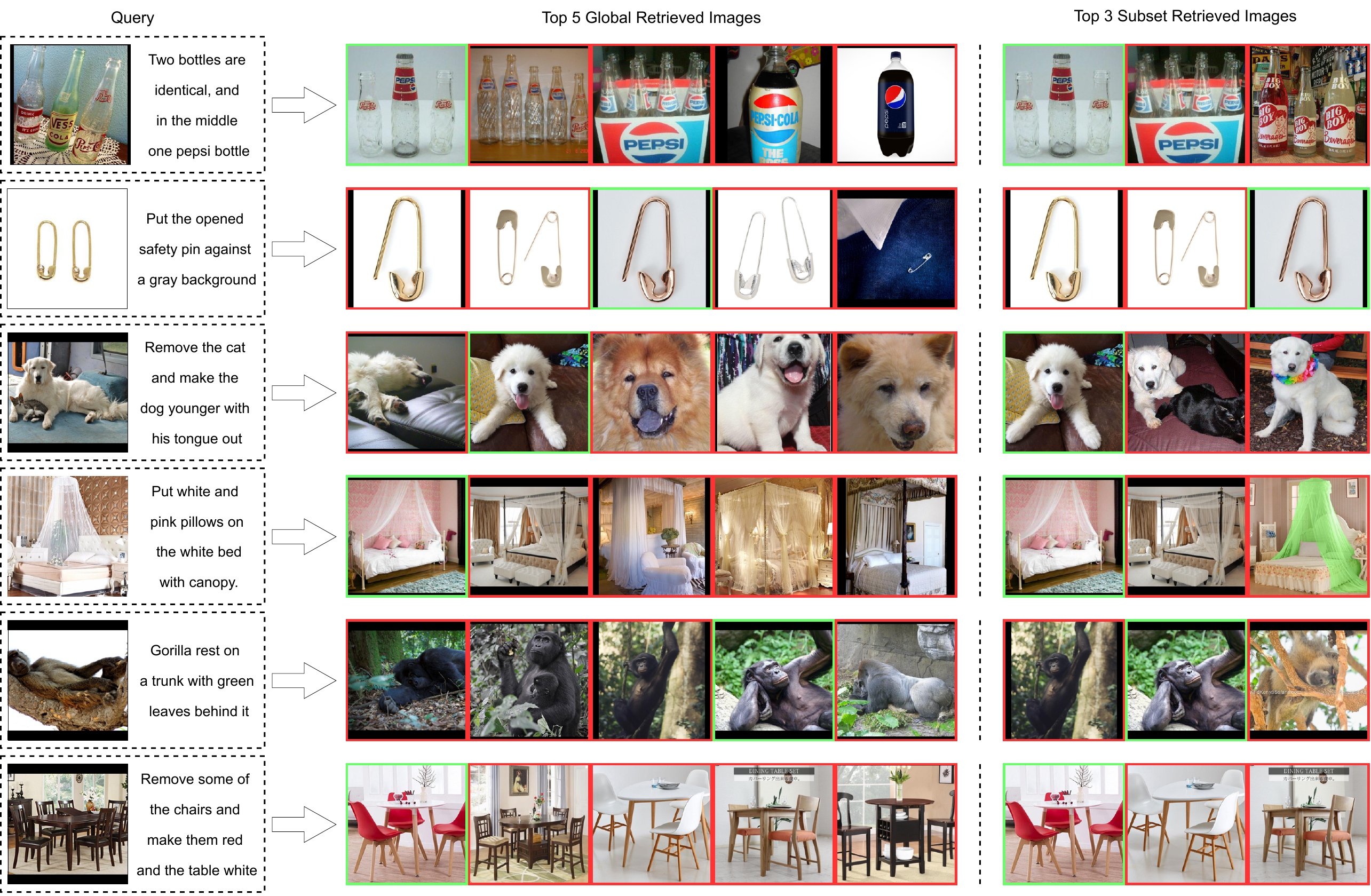}
    \caption{Qualitative results for the CIRR dataset. We highlight with a green border when the retrieved image is labeled as ground truth for the given query.}
    \label{fig:cirr_retrieved}
    \vspace{-2ex}
\end{figure*}

Finally, we complete the qualitative analysis of our approach by presenting examples of multimodal queries and their corresponding results on both datasets in \cref{fig:cirr_retrieved} and \cref{fig:fashion_retrieved}. In FashionIQ, the correct result is returned most of the time in the first three results, while in CIRR, it is returned in the top-5 global and top-3 subset results.
Interestingly, the excellent performance of the proposed system let us notice an issue with the FashionIQ dataset: from these examples, we can see that in the FashionIQ dataset, the existence of many false negatives is a real issue that can harm both the results and the training process; examining the first four and the last queries, we can observe that several results returned in the first positions are corresponding to the conditioning text, although only one of them is marked as such. E.g.~in the first query, where several dresses have light floral patterns and bright colors, similarly, the first three results for the shirts should be considered correct. 
We can also see that in the CIRR dataset, the domain of the images is wider compared to FashionIQ, and the problem of the false negatives is a minor issue. 

\section{Conclusions}\label{sec:conclusion}
In this work, we propose a novel task-oriented fine-tuning scheme to adapt vision-language models for the composed image retrieval task. The primary goal of this fine-tuning is to address the mismatch between the large-scale pre-training of CLIP and the downstream task, thereby enhancing the additivity properties of the embedding spaces.
We then propose a two-stage approach that combines fine-tuning with the training of a carefully crafted Combiner network, enabling the meaningful fusion of the fine-tuned multimodal features. To further enhance performance, we introduce a novel pre-processing padding method, which, as demonstrated in the ablation studies, improves performance on datasets with images of varying aspect ratios. We perform experiments on the challenging fashion dataset FashionIQ and the recently presented CIRR dataset. Experiments on both datasets show that our approach outperforms state-of-the-art methods by a significant margin.
We also perform qualitative experiments to explain how our approach works. These experiments investigate the impact of the proposed approach on the feature distribution in the embedding spaces and how the reshaping of such embedding spaces influences retrieval performance. Additionally, we conduct visualization experiments using the gradCAM technique.
% Our future work will deal with the extension of the proposed method to videos and further experimentation with different image domains, e.g.~cultural heritage.

% %%
% %% The acknowledgments section is defined using the "acks" environment
% %% (and NOT an unnumbered section). This ensures the proper
% %% identification of the section in the article metadata, and the
% %% consistent spelling of the heading.
\begin{acks}
This work was partially supported by the European Commission under European Horizon 2020 Programme, grant number 101004545 - ReInHerit.
\end{acks}
%\balance

\bibliographystyle{ACM-Reference-Format}
\bibliography{CLIP_image_retrieval}


\begin{thebibliography}{56}

%%% the Web address.

\ifx \showCODEN    \undefined \def \showCODEN     #1{\unskip}     \fi
\ifx \showDOI      \undefined \def \showDOI       #1{#1}\fi
\ifx \showISBNx    \undefined \def \showISBNx     #1{\unskip}     \fi
\ifx \showISBNxiii \undefined \def \showISBNxiii  #1{\unskip}     \fi
\ifx \showISSN     \undefined \def \showISSN      #1{\unskip}     \fi
\ifx \showLCCN     \undefined \def \showLCCN      #1{\unskip}     \fi
\ifx \shownote     \undefined \def \shownote      #1{#1}          \fi
\ifx \showarticletitle \undefined \def \showarticletitle #1{#1}   \fi
\ifx \showURL      \undefined \def \showURL       {\relax}        \fi
% invisible to TeX
\providecommand\bibfield[2]{#2}
\providecommand\bibinfo[2]{#2}
\providecommand\natexlab[1]{#1}
\providecommand\showeprint[2][]{arXiv:#2}

\bibitem[\protect\citeauthoryear{Agarwal, Krueger, Clark, Radford, Kim, and
  Brundage}{Agarwal et~al\mbox{.}}{2021}]%
        {agarwal2021evaluating}
\bibfield{author}{\bibinfo{person}{Sandhini Agarwal}, \bibinfo{person}{Gretchen
  Krueger}, \bibinfo{person}{Jack Clark}, \bibinfo{person}{Alec Radford},
  \bibinfo{person}{Jong~Wook Kim}, {and} \bibinfo{person}{Miles Brundage}.}
  \bibinfo{year}{2021}\natexlab{}.
\newblock \showarticletitle{Evaluating clip: towards characterization of
  broader capabilities and downstream implications}.
\newblock \bibinfo{journal}{\emph{arXiv preprint arXiv:2108.02818}}
  (\bibinfo{year}{2021}).
\newblock


\bibitem[\protect\citeauthoryear{Ahmad, Muhammad, Bakshi, and Baik}{Ahmad
  et~al\mbox{.}}{2018}]%
        {ahmad2018object}
\bibfield{author}{\bibinfo{person}{Jamil Ahmad}, \bibinfo{person}{Khan
  Muhammad}, \bibinfo{person}{Sambit Bakshi}, {and} \bibinfo{person}{Sung~Wook
  Baik}.} \bibinfo{year}{2018}\natexlab{}.
\newblock \showarticletitle{Object-oriented convolutional features for
  fine-grained image retrieval in large surveillance datasets}.
\newblock \bibinfo{journal}{\emph{Future Generation Computer Systems}}
  \bibinfo{volume}{81} (\bibinfo{year}{2018}), \bibinfo{pages}{314--330}.
\newblock


\bibitem[\protect\citeauthoryear{Anwaar, Labintcev, and Kleinsteuber}{Anwaar
  et~al\mbox{.}}{2021}]%
        {Anwaar_2021_WACV}
\bibfield{author}{\bibinfo{person}{Muhammad~Umer Anwaar}, \bibinfo{person}{Egor
  Labintcev}, {and} \bibinfo{person}{Martin Kleinsteuber}.}
  \bibinfo{year}{2021}\natexlab{}.
\newblock \showarticletitle{Compositional Learning of Image-Text Query for
  Image Retrieval}. In \bibinfo{booktitle}{\emph{Proc. of IEEE/CVF Winter
  Conference on Applications of Computer Vision (WACV)}}.
  \bibinfo{pages}{1140--1149}.
\newblock


\bibitem[\protect\citeauthoryear{Baldrati, Bertini, Uricchio, and
  Del~Bimbo}{Baldrati et~al\mbox{.}}{2022}]%
        {baldrati2022exploiting}
\bibfield{author}{\bibinfo{person}{Alberto Baldrati}, \bibinfo{person}{Marco
  Bertini}, \bibinfo{person}{Tiberio Uricchio}, {and} \bibinfo{person}{Alberto
  Del~Bimbo}.} \bibinfo{year}{2022}\natexlab{}.
\newblock \showarticletitle{Exploiting CLIP-Based Multi-modal Approach for
  Artwork Classification and Retrieval}. In \bibinfo{booktitle}{\emph{The
  Future of Heritage Science and Technologies: ICT and Digital Heritage: Third
  International Conference, Florence Heri-Tech 2022, Florence, Italy, May
  16--18, 2022, Proceedings}}. Springer, \bibinfo{pages}{140--149}.
\newblock


\bibitem[\protect\citeauthoryear{Banerjee, Kurtz, Devorah, Do, Rubin, and
  Beaulieu}{Banerjee et~al\mbox{.}}{2018}]%
        {banerjee2018relevance}
\bibfield{author}{\bibinfo{person}{Imon Banerjee}, \bibinfo{person}{Camille
  Kurtz}, \bibinfo{person}{Alon~Edward Devorah}, \bibinfo{person}{Bao Do},
  \bibinfo{person}{Daniel~L Rubin}, {and} \bibinfo{person}{Christopher~F
  Beaulieu}.} \bibinfo{year}{2018}\natexlab{}.
\newblock \showarticletitle{Relevance feedback for enhancing content based
  image retrieval and automatic prediction of semantic image features:
  Application to bone tumor radiographs}.
\newblock \bibinfo{journal}{\emph{Journal of biomedical informatics}}
  \bibinfo{volume}{84} (\bibinfo{year}{2018}), \bibinfo{pages}{123--135}.
\newblock


\bibitem[\protect\citeauthoryear{Brown, Mann, Ryder, Subbiah, Kaplan, Dhariwal,
  Neelakantan, Shyam, Sastry, Askell, et~al\mbox{.}}{Brown
  et~al\mbox{.}}{2020}]%
        {brown2020language}
\bibfield{author}{\bibinfo{person}{Tom Brown}, \bibinfo{person}{Benjamin Mann},
  \bibinfo{person}{Nick Ryder}, \bibinfo{person}{Melanie Subbiah},
  \bibinfo{person}{Jared~D Kaplan}, \bibinfo{person}{Prafulla Dhariwal},
  \bibinfo{person}{Arvind Neelakantan}, \bibinfo{person}{Pranav Shyam},
  \bibinfo{person}{Girish Sastry}, \bibinfo{person}{Amanda Askell},
  {et~al\mbox{.}}} \bibinfo{year}{2020}\natexlab{}.
\newblock \showarticletitle{Language models are few-shot learners}.
\newblock \bibinfo{journal}{\emph{Advances in neural information processing
  systems}}  \bibinfo{volume}{33} (\bibinfo{year}{2020}),
  \bibinfo{pages}{1877--1901}.
\newblock


\bibitem[\protect\citeauthoryear{Chen, Xu, Zhang, and Guestrin}{Chen
  et~al\mbox{.}}{2016}]%
        {chen2016training}
\bibfield{author}{\bibinfo{person}{Tianqi Chen}, \bibinfo{person}{Bing Xu},
  \bibinfo{person}{Chiyuan Zhang}, {and} \bibinfo{person}{Carlos Guestrin}.}
  \bibinfo{year}{2016}\natexlab{}.
\newblock \showarticletitle{Training deep nets with sublinear memory cost}.
\newblock \bibinfo{journal}{\emph{arXiv preprint arXiv:1604.06174}}
  (\bibinfo{year}{2016}).
\newblock


\bibitem[\protect\citeauthoryear{Chen, Gong, and Bazzani}{Chen
  et~al\mbox{.}}{2020}]%
        {Chen_2020_CVPR}
\bibfield{author}{\bibinfo{person}{Yanbei Chen}, \bibinfo{person}{Shaogang
  Gong}, {and} \bibinfo{person}{Loris Bazzani}.}
  \bibinfo{year}{2020}\natexlab{}.
\newblock \showarticletitle{Image Search With Text Feedback by Visiolinguistic
  Attention Learning}. In \bibinfo{booktitle}{\emph{Proc. of Conference on
  Computer Vision and Pattern Recognition (CVPR)}}.
\newblock


\bibitem[\protect\citeauthoryear{Cheng, Wu, Zhang, Vajda, and Gonzalez}{Cheng
  et~al\mbox{.}}{2021}]%
        {Cheng_2021_CVPR}
\bibfield{author}{\bibinfo{person}{Ruizhe Cheng}, \bibinfo{person}{Bichen Wu},
  \bibinfo{person}{Peizhao Zhang}, \bibinfo{person}{Peter Vajda}, {and}
  \bibinfo{person}{Joseph~E. Gonzalez}.} \bibinfo{year}{2021}\natexlab{}.
\newblock \showarticletitle{Data-Efficient Language-Supervised Zero-Shot
  Learning With Self-Distillation}. In \bibinfo{booktitle}{\emph{Proc. of
  IEEE/CVF Conference on Computer Vision and Pattern Recognition (CVPR)
  Workshops}}. \bibinfo{pages}{3119--3124}.
\newblock


\bibitem[\protect\citeauthoryear{Chung, Gulcehre, Cho, and Bengio}{Chung
  et~al\mbox{.}}{2014}]%
        {chung2014empirical}
\bibfield{author}{\bibinfo{person}{Junyoung Chung}, \bibinfo{person}{Caglar
  Gulcehre}, \bibinfo{person}{KyungHyun Cho}, {and} \bibinfo{person}{Yoshua
  Bengio}.} \bibinfo{year}{2014}\natexlab{}.
\newblock \showarticletitle{Empirical evaluation of gated recurrent neural
  networks on sequence modeling}.
\newblock \bibinfo{journal}{\emph{arXiv preprint arXiv:1412.3555}}
  (\bibinfo{year}{2014}).
\newblock


\bibitem[\protect\citeauthoryear{Cimino, Galatolo, and Vaglini}{Cimino
  et~al\mbox{.}}{2021}]%
        {cimino2021generating}
\bibfield{author}{\bibinfo{person}{Mario~GCA Cimino},
  \bibinfo{person}{Federico~A Galatolo}, {and} \bibinfo{person}{Gigliola
  Vaglini}.} \bibinfo{year}{2021}\natexlab{}.
\newblock \showarticletitle{Generating Images from Caption and Vice Versa via
  CLIP-Guided Generative Latent Space Search}. In
  \bibinfo{booktitle}{\emph{Proceedings of the International Conference on
  Image Processing and Vision Engineering}}. \bibinfo{pages}{166--174}.
\newblock


\bibitem[\protect\citeauthoryear{Companioni-Brito, Mariano-Calibjio, Elawady,
  and Yildirim}{Companioni-Brito et~al\mbox{.}}{2018}]%
        {yildirim2018mobile}
\bibfield{author}{\bibinfo{person}{Claudia Companioni-Brito},
  \bibinfo{person}{Zygred Mariano-Calibjio}, \bibinfo{person}{Mohamed Elawady},
  {and} \bibinfo{person}{Sule Yildirim}.} \bibinfo{year}{2018}\natexlab{}.
\newblock \showarticletitle{Mobile-Based Painting Photo Retrieval Using
  Combined Features}. In \bibinfo{booktitle}{\emph{Proc. of International
  Conference on Image Analysis and Recognition (ICIAR)}},
  Vol.~\bibinfo{volume}{10882}. Springer, \bibinfo{pages}{278}.
\newblock


\bibitem[\protect\citeauthoryear{Conde and Turgutlu}{Conde and
  Turgutlu}{2021}]%
        {conde2021clip}
\bibfield{author}{\bibinfo{person}{Marcos~V Conde} {and} \bibinfo{person}{Kerem
  Turgutlu}.} \bibinfo{year}{2021}\natexlab{}.
\newblock \showarticletitle{{CLIP-Art}: Contrastive Pre-Training for
  Fine-Grained Art Classification}. In \bibinfo{booktitle}{\emph{Proc. of
  Conference on Computer Vision and Pattern Recognition (CVPR)}}.
  \bibinfo{pages}{3956--3960}.
\newblock


\bibitem[\protect\citeauthoryear{Delmas, Rezende, Csurka, and Larlus}{Delmas
  et~al\mbox{.}}{2021}]%
        {delmas2021artemis}
\bibfield{author}{\bibinfo{person}{Ginger Delmas}, \bibinfo{person}{Rafael~S
  Rezende}, \bibinfo{person}{Gabriela Csurka}, {and} \bibinfo{person}{Diane
  Larlus}.} \bibinfo{year}{2021}\natexlab{}.
\newblock \showarticletitle{ARTEMIS: Attention-based Retrieval with
  Text-Explicit Matching and Implicit Similarity}. In
  \bibinfo{booktitle}{\emph{International Conference on Learning
  Representations}}.
\newblock


\bibitem[\protect\citeauthoryear{Devlin, Chang, Lee, and Toutanova}{Devlin
  et~al\mbox{.}}{2019}]%
        {devlin2019bert}
\bibfield{author}{\bibinfo{person}{Jacob Devlin}, \bibinfo{person}{Ming-Wei
  Chang}, \bibinfo{person}{Kenton Lee}, {and} \bibinfo{person}{Kristina
  Toutanova}.} \bibinfo{year}{2019}\natexlab{}.
\newblock \showarticletitle{BERT: Pre-training of Deep Bidirectional
  Transformers for Language Understanding}. In
  \bibinfo{booktitle}{\emph{Proceedings of the 2019 Conference of the North
  American Chapter of the Association for Computational Linguistics: Human
  Language Technologies, Volume 1 (Long and Short Papers)}}.
  \bibinfo{pages}{4171--4186}.
\newblock


\bibitem[\protect\citeauthoryear{Dodds, Culpepper, Herdade, Zhang, and
  Boakye}{Dodds et~al\mbox{.}}{2020}]%
        {dodds2020modality}
\bibfield{author}{\bibinfo{person}{Eric Dodds}, \bibinfo{person}{Jack
  Culpepper}, \bibinfo{person}{Simao Herdade}, \bibinfo{person}{Yang Zhang},
  {and} \bibinfo{person}{Kofi Boakye}.} \bibinfo{year}{2020}\natexlab{}.
\newblock \showarticletitle{Modality-agnostic attention fusion for visual
  search with text feedback}.
\newblock \bibinfo{journal}{\emph{arXiv preprint arXiv:2007.00145}}
  (\bibinfo{year}{2020}).
\newblock


\bibitem[\protect\citeauthoryear{Dong, Zhan, Wu, Wei, Wei, Lu, and Liang}{Dong
  et~al\mbox{.}}{2021}]%
        {dong2021m5product}
\bibfield{author}{\bibinfo{person}{Xiao Dong}, \bibinfo{person}{Xunlin Zhan},
  \bibinfo{person}{Yangxin Wu}, \bibinfo{person}{Yunchao Wei},
  \bibinfo{person}{Xiaoyong Wei}, \bibinfo{person}{Minlong Lu}, {and}
  \bibinfo{person}{Xiaodan Liang}.} \bibinfo{year}{2021}\natexlab{}.
\newblock \showarticletitle{M5product: A multi-modal pretraining benchmark for
  e-commercial product downstream tasks}.
\newblock \bibinfo{journal}{\emph{arXiv preprint arXiv:2109.04275}}
  (\bibinfo{year}{2021}).
\newblock


\bibitem[\protect\citeauthoryear{Fang, Xiong, Xu, and Chen}{Fang
  et~al\mbox{.}}{2021}]%
        {fang2021clip2video}
\bibfield{author}{\bibinfo{person}{Han Fang}, \bibinfo{person}{Pengfei Xiong},
  \bibinfo{person}{Luhui Xu}, {and} \bibinfo{person}{Yu Chen}.}
  \bibinfo{year}{2021}\natexlab{}.
\newblock \showarticletitle{{CLIP2Video}: Mastering Video-Text Retrieval via
  Image {CLIP}}.
\newblock \bibinfo{journal}{\emph{arXiv preprint arXiv:2106.11097}}
  (\bibinfo{year}{2021}).
\newblock


\bibitem[\protect\citeauthoryear{Guo, Wu, Cheng, Rennie, Tesauro, and
  Feris}{Guo et~al\mbox{.}}{2018}]%
        {guo2018dialog}
\bibfield{author}{\bibinfo{person}{Xiaoxiao Guo}, \bibinfo{person}{Hui Wu},
  \bibinfo{person}{Yu Cheng}, \bibinfo{person}{Steven Rennie},
  \bibinfo{person}{Gerald Tesauro}, {and} \bibinfo{person}{Rogerio Feris}.}
  \bibinfo{year}{2018}\natexlab{}.
\newblock \showarticletitle{Dialog-based interactive image retrieval}.
\newblock \bibinfo{journal}{\emph{Advances in neural information processing
  systems}}  \bibinfo{volume}{31} (\bibinfo{year}{2018}).
\newblock


\bibitem[\protect\citeauthoryear{Han, Wu, Huang, Zhang, Zhu, Li, Zhao, and
  Davis}{Han et~al\mbox{.}}{2017}]%
        {han2017automatic}
\bibfield{author}{\bibinfo{person}{Xintong Han}, \bibinfo{person}{Zuxuan Wu},
  \bibinfo{person}{Phoenix~X Huang}, \bibinfo{person}{Xiao Zhang},
  \bibinfo{person}{Menglong Zhu}, \bibinfo{person}{Yuan Li},
  \bibinfo{person}{Yang Zhao}, {and} \bibinfo{person}{Larry~S Davis}.}
  \bibinfo{year}{2017}\natexlab{}.
\newblock \showarticletitle{Automatic spatially-aware fashion concept
  discovery}. In \bibinfo{booktitle}{\emph{Proceedings of the IEEE
  international conference on computer vision}}. \bibinfo{pages}{1463--1471}.
\newblock


\bibitem[\protect\citeauthoryear{Han, Yu, Zhu, Zhang, Song, and Xiang}{Han
  et~al\mbox{.}}{2022}]%
        {han2022fashionvil}
\bibfield{author}{\bibinfo{person}{Xiao Han}, \bibinfo{person}{Licheng Yu},
  \bibinfo{person}{Xiatian Zhu}, \bibinfo{person}{Li Zhang},
  \bibinfo{person}{Yi-Zhe Song}, {and} \bibinfo{person}{Tao Xiang}.}
  \bibinfo{year}{2022}\natexlab{}.
\newblock \showarticletitle{Fashionvil: Fashion-focused vision-and-language
  representation learning}. In \bibinfo{booktitle}{\emph{European Conference on
  Computer Vision}}. Springer, \bibinfo{pages}{634--651}.
\newblock


\bibitem[\protect\citeauthoryear{He, Zhang, Ren, and Sun}{He
  et~al\mbox{.}}{2016}]%
        {he2016deep}
\bibfield{author}{\bibinfo{person}{Kaiming He}, \bibinfo{person}{Xiangyu
  Zhang}, \bibinfo{person}{Shaoqing Ren}, {and} \bibinfo{person}{Jian Sun}.}
  \bibinfo{year}{2016}\natexlab{}.
\newblock \showarticletitle{Deep residual learning for image recognition}. In
  \bibinfo{booktitle}{\emph{Proceedings of the IEEE conference on computer
  vision and pattern recognition}}. \bibinfo{pages}{770--778}.
\newblock


\bibitem[\protect\citeauthoryear{Hochreiter and Schmidhuber}{Hochreiter and
  Schmidhuber}{1997}]%
        {hochreiter1997long}
\bibfield{author}{\bibinfo{person}{Sepp Hochreiter} {and}
  \bibinfo{person}{J{\"u}rgen Schmidhuber}.} \bibinfo{year}{1997}\natexlab{}.
\newblock \showarticletitle{Long short-term memory}.
\newblock \bibinfo{journal}{\emph{Neural computation}} \bibinfo{volume}{9},
  \bibinfo{number}{8} (\bibinfo{year}{1997}), \bibinfo{pages}{1735--1780}.
\newblock


\bibitem[\protect\citeauthoryear{Ionescu, M{\"u}ller, P{\'e}teri, Cid,
  Liauchuk, Kovalev, Klimuk, Tarasau, Abacha, Hasan, et~al\mbox{.}}{Ionescu
  et~al\mbox{.}}{2019}]%
        {ionescu2019imageclef}
\bibfield{author}{\bibinfo{person}{Bogdan Ionescu}, \bibinfo{person}{Henning
  M{\"u}ller}, \bibinfo{person}{Renaud P{\'e}teri},
  \bibinfo{person}{Yashin~Dicente Cid}, \bibinfo{person}{Vitali Liauchuk},
  \bibinfo{person}{Vassili Kovalev}, \bibinfo{person}{Dzmitri Klimuk},
  \bibinfo{person}{Aleh Tarasau}, \bibinfo{person}{Asma~Ben Abacha},
  \bibinfo{person}{Sadid~A Hasan}, {et~al\mbox{.}}}
  \bibinfo{year}{2019}\natexlab{}.
\newblock \showarticletitle{{ImageCLEF} 2019: Multimedia retrieval in medicine,
  lifelogging, security and nature}. In \bibinfo{booktitle}{\emph{Proc. of
  International Conference of the Cross-Language Evaluation Forum for European
  Languages (CLEF)}}. Springer, \bibinfo{pages}{358--386}.
\newblock


\bibitem[\protect\citeauthoryear{Ionescu, M{\"u}ller, P{\'e}teri, Dang-Nguyen,
  Zhou, Piras, Riegler, Halvorsen, Tran, Lux, et~al\mbox{.}}{Ionescu
  et~al\mbox{.}}{2020}]%
        {ionescu2020imageclef}
\bibfield{author}{\bibinfo{person}{Bogdan Ionescu}, \bibinfo{person}{Henning
  M{\"u}ller}, \bibinfo{person}{Renaud P{\'e}teri}, \bibinfo{person}{Duc-Tien
  Dang-Nguyen}, \bibinfo{person}{Liting Zhou}, \bibinfo{person}{Luca Piras},
  \bibinfo{person}{Michael Riegler}, \bibinfo{person}{P{\aa}l Halvorsen},
  \bibinfo{person}{Minh-Triet Tran}, \bibinfo{person}{Mathias Lux},
  {et~al\mbox{.}}} \bibinfo{year}{2020}\natexlab{}.
\newblock \showarticletitle{{ImageCLEF} 2020: Multimedia retrieval in
  lifelogging, medical, nature, and internet applications}.
\newblock \bibinfo{journal}{\emph{Advances in Information Retrieval}}
  \bibinfo{volume}{12036} (\bibinfo{year}{2020}), \bibinfo{pages}{533}.
\newblock


\bibitem[\protect\citeauthoryear{Jandial, Badjatiya, Chawla, Chopra, Sarkar,
  and Krishnamurthy}{Jandial et~al\mbox{.}}{2022}]%
        {Jandial_2022_WACV}
\bibfield{author}{\bibinfo{person}{Surgan Jandial}, \bibinfo{person}{Pinkesh
  Badjatiya}, \bibinfo{person}{Pranit Chawla}, \bibinfo{person}{Ayush Chopra},
  \bibinfo{person}{Mausoom Sarkar}, {and} \bibinfo{person}{Balaji
  Krishnamurthy}.} \bibinfo{year}{2022}\natexlab{}.
\newblock \showarticletitle{{SAC}: Semantic Attention Composition for
  Text-Conditioned Image Retrieval}. In \bibinfo{booktitle}{\emph{Proc. of
  IEEE/CVF Winter Conference on Applications of Computer Vision (WACV)}}.
  \bibinfo{pages}{4021--4030}.
\newblock


\bibitem[\protect\citeauthoryear{Jandial, Chopra, Badjatiya, Chawla, Sarkar,
  and Krishnamurthy}{Jandial et~al\mbox{.}}{2020}]%
        {jandial2020trace}
\bibfield{author}{\bibinfo{person}{Surgan Jandial}, \bibinfo{person}{Ayush
  Chopra}, \bibinfo{person}{Pinkesh Badjatiya}, \bibinfo{person}{Pranit
  Chawla}, \bibinfo{person}{Mausoom Sarkar}, {and} \bibinfo{person}{Balaji
  Krishnamurthy}.} \bibinfo{year}{2020}\natexlab{}.
\newblock \showarticletitle{Trace: Transform aggregate and compose
  visiolinguistic representations for image search with text feedback}.
\newblock \bibinfo{journal}{\emph{arXiv preprint arXiv:2009.01485}}
  \bibinfo{volume}{7} (\bibinfo{year}{2020}), \bibinfo{pages}{7}.
\newblock


\bibitem[\protect\citeauthoryear{Jia, Yang, Xia, Chen, Parekh, Pham, Le, Sung,
  Li, and Duerig}{Jia et~al\mbox{.}}{2021}]%
        {jia2021scaling}
\bibfield{author}{\bibinfo{person}{Chao Jia}, \bibinfo{person}{Yinfei Yang},
  \bibinfo{person}{Ye Xia}, \bibinfo{person}{Yi-Ting Chen},
  \bibinfo{person}{Zarana Parekh}, \bibinfo{person}{Hieu Pham},
  \bibinfo{person}{Quoc~V Le}, \bibinfo{person}{Yunhsuan Sung},
  \bibinfo{person}{Zhen Li}, {and} \bibinfo{person}{Tom Duerig}.}
  \bibinfo{year}{2021}\natexlab{}.
\newblock \showarticletitle{Scaling up visual and vision-language
  representation learning with noisy text supervision}. In
  \bibinfo{booktitle}{\emph{Proc. of International Conference on Machine
  Learning (ICML)}}.
\newblock


\bibitem[\protect\citeauthoryear{Kim, Yu, Kim, and Kim}{Kim
  et~al\mbox{.}}{2021}]%
        {Kim_Yu_Kim_Kim_2021}
\bibfield{author}{\bibinfo{person}{Jongseok Kim}, \bibinfo{person}{Youngjae
  Yu}, \bibinfo{person}{Hoeseong Kim}, {and} \bibinfo{person}{Gunhee Kim}.}
  \bibinfo{year}{2021}\natexlab{}.
\newblock \showarticletitle{Dual Compositional Learning in Interactive Image
  Retrieval}. In \bibinfo{booktitle}{\emph{Proc. of AAAI Conference on
  Artificial Intelligence (AAAI)}}, Vol.~\bibinfo{volume}{35}.
  \bibinfo{pages}{1771--1779}.
\newblock
\urldef\tempurl%
\url{https://ojs.aaai.org/index.php/AAAI/article/view/16271}
\showURL{%
\tempurl}


\bibitem[\protect\citeauthoryear{Kovashka, Parikh, and Grauman}{Kovashka
  et~al\mbox{.}}{2015}]%
        {Kovashka_2015}
\bibfield{author}{\bibinfo{person}{Adriana Kovashka}, \bibinfo{person}{Devi
  Parikh}, {and} \bibinfo{person}{Kristen Grauman}.}
  \bibinfo{year}{2015}\natexlab{}.
\newblock \showarticletitle{WhittleSearch: Interactive Image Search with
  Relative Attribute Feedback}.
\newblock \bibinfo{journal}{\emph{International Journal of Computer Vision
  (IJCV)}} \bibinfo{volume}{115}, \bibinfo{number}{2} (\bibinfo{date}{Apr}
  \bibinfo{year}{2015}), \bibinfo{pages}{185--210}.
\newblock
\showISSN{1573-1405}
\urldef\tempurl%
\url{https://doi.org/10.1007/s11263-015-0814-0}
\showDOI{\tempurl}


\bibitem[\protect\citeauthoryear{Lee, Kim, and Han}{Lee et~al\mbox{.}}{2021}]%
        {Lee_2021_CVPR}
\bibfield{author}{\bibinfo{person}{Seungmin Lee}, \bibinfo{person}{Dongwan
  Kim}, {and} \bibinfo{person}{Bohyung Han}.} \bibinfo{year}{2021}\natexlab{}.
\newblock \showarticletitle{{CoSMo}: Content-Style Modulation for Image
  Retrieval With Text Feedback}. In \bibinfo{booktitle}{\emph{Proc. of
  Conference on Computer Vision and Pattern Recognition (CVPR)}}.
  \bibinfo{pages}{802--812}.
\newblock


\bibitem[\protect\citeauthoryear{Li, Xu, Wang, Zhou, Lin, Zhu, Zeng, Ji, and
  Chang}{Li et~al\mbox{.}}{2022}]%
        {li2022clip}
\bibfield{author}{\bibinfo{person}{Manling Li}, \bibinfo{person}{Ruochen Xu},
  \bibinfo{person}{Shuohang Wang}, \bibinfo{person}{Luowei Zhou},
  \bibinfo{person}{Xudong Lin}, \bibinfo{person}{Chenguang Zhu},
  \bibinfo{person}{Michael Zeng}, \bibinfo{person}{Heng Ji}, {and}
  \bibinfo{person}{Shih-Fu Chang}.} \bibinfo{year}{2022}\natexlab{}.
\newblock \showarticletitle{Clip-event: Connecting text and images with event
  structures}. In \bibinfo{booktitle}{\emph{Proceedings of the IEEE/CVF
  Conference on Computer Vision and Pattern Recognition}}.
  \bibinfo{pages}{16420--16429}.
\newblock


\bibitem[\protect\citeauthoryear{Li, Yang, and Ma}{Li et~al\mbox{.}}{2021}]%
        {LI2021675}
\bibfield{author}{\bibinfo{person}{Xiaoqing Li}, \bibinfo{person}{Jiansheng
  Yang}, {and} \bibinfo{person}{Jinwen Ma}.} \bibinfo{year}{2021}\natexlab{}.
\newblock \showarticletitle{Recent developments of content-based image
  retrieval ({CBIR})}.
\newblock \bibinfo{journal}{\emph{Neurocomputing}}  \bibinfo{volume}{452}
  (\bibinfo{year}{2021}), \bibinfo{pages}{675--689}.
\newblock
\showISSN{0925-2312}
\urldef\tempurl%
\url{https://doi.org/10.1016/j.neucom.2020.07.139}
\showDOI{\tempurl}


\bibitem[\protect\citeauthoryear{Li, Yin, Li, Zhang, Hu, Zhang, Wang, Hu, Dong,
  Wei, et~al\mbox{.}}{Li et~al\mbox{.}}{2020}]%
        {li2020oscar}
\bibfield{author}{\bibinfo{person}{Xiujun Li}, \bibinfo{person}{Xi Yin},
  \bibinfo{person}{Chunyuan Li}, \bibinfo{person}{Pengchuan Zhang},
  \bibinfo{person}{Xiaowei Hu}, \bibinfo{person}{Lei Zhang},
  \bibinfo{person}{Lijuan Wang}, \bibinfo{person}{Houdong Hu},
  \bibinfo{person}{Li Dong}, \bibinfo{person}{Furu Wei}, {et~al\mbox{.}}}
  \bibinfo{year}{2020}\natexlab{}.
\newblock \showarticletitle{Oscar: Object-semantics aligned pre-training for
  vision-language tasks}. In \bibinfo{booktitle}{\emph{Computer Vision--ECCV
  2020: 16th European Conference, Glasgow, UK, August 23--28, 2020,
  Proceedings, Part XXX 16}}. Springer, \bibinfo{pages}{121--137}.
\newblock


\bibitem[\protect\citeauthoryear{Liang, Zhang, Kwon, Yeung, and Zou}{Liang
  et~al\mbox{.}}{2022}]%
        {liang2022mind}
\bibfield{author}{\bibinfo{person}{Victor~Weixin Liang}, \bibinfo{person}{Yuhui
  Zhang}, \bibinfo{person}{Yongchan Kwon}, \bibinfo{person}{Serena Yeung},
  {and} \bibinfo{person}{James~Y Zou}.} \bibinfo{year}{2022}\natexlab{}.
\newblock \showarticletitle{Mind the gap: Understanding the modality gap in
  multi-modal contrastive representation learning}.
\newblock \bibinfo{journal}{\emph{Advances in Neural Information Processing
  Systems}}  \bibinfo{volume}{35} (\bibinfo{year}{2022}),
  \bibinfo{pages}{17612--17625}.
\newblock


\bibitem[\protect\citeauthoryear{Liu and Lu}{Liu and Lu}{2021}]%
        {Liu-2021-multigrained}
\bibfield{author}{\bibinfo{person}{Yating Liu} {and} \bibinfo{person}{Yan Lu}.}
  \bibinfo{year}{2021}\natexlab{}.
\newblock \showarticletitle{Multi-grained Fusion for Conditional Image
  Retrieval}. In \bibinfo{booktitle}{\emph{Proc. of International Conference on
  Multimedia Modeling (MMM)}}. \bibinfo{publisher}{Springer International
  Publishing}, \bibinfo{address}{Cham}, \bibinfo{pages}{315--327}.
\newblock
\showISBNx{978-3-030-67832-6}


\bibitem[\protect\citeauthoryear{Liu, Rodriguez-Opazo, Teney, and Gould}{Liu
  et~al\mbox{.}}{2021}]%
        {liu2021image}
\bibfield{author}{\bibinfo{person}{Zheyuan Liu}, \bibinfo{person}{Cristian
  Rodriguez-Opazo}, \bibinfo{person}{Damien Teney}, {and}
  \bibinfo{person}{Stephen Gould}.} \bibinfo{year}{2021}\natexlab{}.
\newblock \showarticletitle{Image retrieval on real-life images with
  pre-trained vision-and-language models}. In
  \bibinfo{booktitle}{\emph{Proceedings of the IEEE/CVF International
  Conference on Computer Vision}}. \bibinfo{pages}{2125--2134}.
\newblock


\bibitem[\protect\citeauthoryear{Loshchilov and Hutter}{Loshchilov and
  Hutter}{2019}]%
        {loshchilov2018decoupled}
\bibfield{author}{\bibinfo{person}{Ilya Loshchilov} {and}
  \bibinfo{person}{Frank Hutter}.} \bibinfo{year}{2019}\natexlab{}.
\newblock \showarticletitle{Decoupled Weight Decay Regularization}. In
  \bibinfo{booktitle}{\emph{International Conference on Learning
  Representations}}.
\newblock
\urldef\tempurl%
\url{https://openreview.net/forum?id=Bkg6RiCqY7}
\showURL{%
\tempurl}


\bibitem[\protect\citeauthoryear{Micikevicius, Narang, Alben, Diamos, Elsen,
  Garcia, Ginsburg, Houston, Kuchaiev, Venkatesh, et~al\mbox{.}}{Micikevicius
  et~al\mbox{.}}{2018}]%
        {micikevicius2018mixed}
\bibfield{author}{\bibinfo{person}{Paulius Micikevicius},
  \bibinfo{person}{Sharan Narang}, \bibinfo{person}{Jonah Alben},
  \bibinfo{person}{Gregory Diamos}, \bibinfo{person}{Erich Elsen},
  \bibinfo{person}{David Garcia}, \bibinfo{person}{Boris Ginsburg},
  \bibinfo{person}{Michael Houston}, \bibinfo{person}{Oleksii Kuchaiev},
  \bibinfo{person}{Ganesh Venkatesh}, {et~al\mbox{.}}}
  \bibinfo{year}{2018}\natexlab{}.
\newblock \showarticletitle{Mixed Precision Training}. In
  \bibinfo{booktitle}{\emph{International Conference on Learning
  Representations}}.
\newblock


\bibitem[\protect\citeauthoryear{Pennington, Socher, and Manning}{Pennington
  et~al\mbox{.}}{2014}]%
        {pennington2014glove}
\bibfield{author}{\bibinfo{person}{Jeffrey Pennington},
  \bibinfo{person}{Richard Socher}, {and} \bibinfo{person}{Christopher~D
  Manning}.} \bibinfo{year}{2014}\natexlab{}.
\newblock \showarticletitle{Glove: Global vectors for word representation}. In
  \bibinfo{booktitle}{\emph{Proceedings of the 2014 conference on empirical
  methods in natural language processing (EMNLP)}}.
  \bibinfo{pages}{1532--1543}.
\newblock


\bibitem[\protect\citeauthoryear{Radford, Kim, Hallacy, Ramesh, Goh, Agarwal,
  Sastry, Askell, Mishkin, Clark, et~al\mbox{.}}{Radford et~al\mbox{.}}{2021}]%
        {radford2021learning}
\bibfield{author}{\bibinfo{person}{Alec Radford}, \bibinfo{person}{Jong~Wook
  Kim}, \bibinfo{person}{Chris Hallacy}, \bibinfo{person}{Aditya Ramesh},
  \bibinfo{person}{Gabriel Goh}, \bibinfo{person}{Sandhini Agarwal},
  \bibinfo{person}{Girish Sastry}, \bibinfo{person}{Amanda Askell},
  \bibinfo{person}{Pamela Mishkin}, \bibinfo{person}{Jack Clark},
  {et~al\mbox{.}}} \bibinfo{year}{2021}\natexlab{}.
\newblock \showarticletitle{Learning transferable visual models from natural
  language supervision}. In \bibinfo{booktitle}{\emph{International conference
  on machine learning}}. PMLR, \bibinfo{pages}{8748--8763}.
\newblock


\bibitem[\protect\citeauthoryear{Rui, Huang, Ortega, and Mehrotra}{Rui
  et~al\mbox{.}}{1998}]%
        {718510}
\bibfield{author}{\bibinfo{person}{Yong Rui}, \bibinfo{person}{T.S. Huang},
  \bibinfo{person}{M. Ortega}, {and} \bibinfo{person}{S. Mehrotra}.}
  \bibinfo{year}{1998}\natexlab{}.
\newblock \showarticletitle{Relevance feedback: a power tool for interactive
  content-based image retrieval}.
\newblock \bibinfo{journal}{\emph{IEEE Transactions on Circuits and Systems for
  Video Technology (TCSVT)}} \bibinfo{volume}{8}, \bibinfo{number}{5}
  (\bibinfo{year}{1998}), \bibinfo{pages}{644--655}.
\newblock
\urldef\tempurl%
\url{https://doi.org/10.1109/76.718510}
\showDOI{\tempurl}


\bibitem[\protect\citeauthoryear{Russakovsky, Deng, Su, Krause, Satheesh, Ma,
  Huang, Karpathy, Khosla, Bernstein, et~al\mbox{.}}{Russakovsky
  et~al\mbox{.}}{2015}]%
        {russakovsky2015imagenet}
\bibfield{author}{\bibinfo{person}{Olga Russakovsky}, \bibinfo{person}{Jia
  Deng}, \bibinfo{person}{Hao Su}, \bibinfo{person}{Jonathan Krause},
  \bibinfo{person}{Sanjeev Satheesh}, \bibinfo{person}{Sean Ma},
  \bibinfo{person}{Zhiheng Huang}, \bibinfo{person}{Andrej Karpathy},
  \bibinfo{person}{Aditya Khosla}, \bibinfo{person}{Michael Bernstein},
  {et~al\mbox{.}}} \bibinfo{year}{2015}\natexlab{}.
\newblock \showarticletitle{Imagenet large scale visual recognition challenge}.
\newblock \bibinfo{journal}{\emph{International journal of computer vision}}
  \bibinfo{volume}{115} (\bibinfo{year}{2015}), \bibinfo{pages}{211--252}.
\newblock


\bibitem[\protect\citeauthoryear{Selvaraju, Cogswell, Das, Vedantam, Parikh,
  and Batra}{Selvaraju et~al\mbox{.}}{2019}]%
        {Selvaraju2019gradCAM}
\bibfield{author}{\bibinfo{person}{Ramprasaath~R. Selvaraju},
  \bibinfo{person}{Michael Cogswell}, \bibinfo{person}{Abhishek Das},
  \bibinfo{person}{Ramakrishna Vedantam}, \bibinfo{person}{Devi Parikh}, {and}
  \bibinfo{person}{Dhruv Batra}.} \bibinfo{year}{2019}\natexlab{}.
\newblock \showarticletitle{{Grad-CAM}: Visual Explanations from Deep Networks
  via Gradient-Based Localization}.
\newblock \bibinfo{journal}{\emph{International Journal of Computer Vision
  (IJCV)}} \bibinfo{volume}{128}, \bibinfo{number}{2} (\bibinfo{date}{Oct}
  \bibinfo{year}{2019}), \bibinfo{pages}{336--359}.
\newblock
\showISSN{1573-1405}
\urldef\tempurl%
\url{https://doi.org/10.1007/s11263-019-01228-7}
\showDOI{\tempurl}


\bibitem[\protect\citeauthoryear{Shin, Cho, Ko, and Gu}{Shin
  et~al\mbox{.}}{2021}]%
        {shin2021rtic}
\bibfield{author}{\bibinfo{person}{Minchul Shin}, \bibinfo{person}{Yoonjae
  Cho}, \bibinfo{person}{Byungsoo Ko}, {and} \bibinfo{person}{Geonmo Gu}.}
  \bibinfo{year}{2021}\natexlab{}.
\newblock \showarticletitle{{RTIC}: Residual Learning for Text and Image
  Composition using Graph Convolutional Network}.
\newblock \bibinfo{journal}{\emph{arXiv preprint arXiv:2104.03015}}
  (\bibinfo{year}{2021}).
\newblock


\bibitem[\protect\citeauthoryear{Smeulders, Worring, Santini, Gupta, and
  Jain}{Smeulders et~al\mbox{.}}{2000}]%
        {smeulders2000content}
\bibfield{author}{\bibinfo{person}{Arnold~WM Smeulders},
  \bibinfo{person}{Marcel Worring}, \bibinfo{person}{Simone Santini},
  \bibinfo{person}{Amarnath Gupta}, {and} \bibinfo{person}{Ramesh Jain}.}
  \bibinfo{year}{2000}\natexlab{}.
\newblock \showarticletitle{Content-based image retrieval at the end of the
  early years}.
\newblock \bibinfo{journal}{\emph{IEEE Transactions on Pattern Analysis and
  Machine Intelligence (TPAMI)}} \bibinfo{volume}{22}, \bibinfo{number}{12}
  (\bibinfo{year}{2000}), \bibinfo{pages}{1349--1380}.
\newblock


\bibitem[\protect\citeauthoryear{Suhr, Zhou, Zhang, Zhang, Bai, and Artzi}{Suhr
  et~al\mbox{.}}{2019}]%
        {suhr2019corpus}
\bibfield{author}{\bibinfo{person}{Alane Suhr}, \bibinfo{person}{Stephanie
  Zhou}, \bibinfo{person}{Ally Zhang}, \bibinfo{person}{Iris Zhang},
  \bibinfo{person}{Huajun Bai}, {and} \bibinfo{person}{Yoav Artzi}.}
  \bibinfo{year}{2019}\natexlab{}.
\newblock \showarticletitle{A Corpus for Reasoning about Natural Language
  Grounded in Photographs}. In \bibinfo{booktitle}{\emph{Proceedings of the
  57th Annual Meeting of the Association for Computational Linguistics}}.
  \bibinfo{pages}{6418--6428}.
\newblock


\bibitem[\protect\citeauthoryear{Vo, Jiang, Sun, Murphy, Li, Fei-Fei, and
  Hays}{Vo et~al\mbox{.}}{2019}]%
        {vo2019composing}
\bibfield{author}{\bibinfo{person}{Nam Vo}, \bibinfo{person}{Lu Jiang},
  \bibinfo{person}{Chen Sun}, \bibinfo{person}{Kevin Murphy},
  \bibinfo{person}{Li-Jia Li}, \bibinfo{person}{Li Fei-Fei}, {and}
  \bibinfo{person}{James Hays}.} \bibinfo{year}{2019}\natexlab{}.
\newblock \showarticletitle{Composing text and image for image retrieval-an
  empirical odyssey}. In \bibinfo{booktitle}{\emph{Proceedings of the IEEE/CVF
  conference on computer vision and pattern recognition}}.
  \bibinfo{pages}{6439--6448}.
\newblock


\bibitem[\protect\citeauthoryear{Wang, Codella, Chen, Zhou, Yang, Dai, Xiao,
  You, Chang, and Yuan}{Wang et~al\mbox{.}}{2022}]%
        {wang2022cliptd}
\bibfield{author}{\bibinfo{person}{Zhecan Wang}, \bibinfo{person}{Noel
  Codella}, \bibinfo{person}{Yen-Chun Chen}, \bibinfo{person}{Luowei Zhou},
  \bibinfo{person}{Jianwei Yang}, \bibinfo{person}{Xiyang Dai},
  \bibinfo{person}{Bin Xiao}, \bibinfo{person}{Haoxuan You},
  \bibinfo{person}{Shih-Fu Chang}, {and} \bibinfo{person}{Lu Yuan}.}
  \bibinfo{year}{2022}\natexlab{}.
\newblock \showarticletitle{{CLIP-TD}: {CLIP} Targeted Distillation for
  Vision-Language Tasks}.
\newblock \bibinfo{journal}{\emph{arXiv preprint arXiv:2201.05729}}
  (\bibinfo{year}{2022}).
\newblock
\showeprint[arxiv]{2201.05729}~[cs.CV]


\bibitem[\protect\citeauthoryear{Wen, Song, Yang, Zhan, and Nie}{Wen
  et~al\mbox{.}}{2021}]%
        {clvc-net}
\bibfield{author}{\bibinfo{person}{Haokun Wen}, \bibinfo{person}{Xuemeng Song},
  \bibinfo{person}{Xin Yang}, \bibinfo{person}{Yibing Zhan}, {and}
  \bibinfo{person}{Liqiang Nie}.} \bibinfo{year}{2021}\natexlab{}.
\newblock \showarticletitle{Comprehensive Linguistic-Visual Composition Network
  for Image Retrieval}. In \bibinfo{booktitle}{\emph{Proceedings of the 44th
  International ACM SIGIR Conference on Research and Development in Information
  Retrieval}} (Virtual Event, Canada) \emph{(\bibinfo{series}{SIGIR '21})}.
  \bibinfo{publisher}{Association for Computing Machinery},
  \bibinfo{address}{New York, NY, USA}, \bibinfo{pages}{1369–1378}.
\newblock
\showISBNx{9781450380379}
\urldef\tempurl%
\url{https://doi.org/10.1145/3404835.3462967}
\showDOI{\tempurl}


\bibitem[\protect\citeauthoryear{Wu, Gao, Guo, Al-Halah, Rennie, Grauman, and
  Feris}{Wu et~al\mbox{.}}{2021}]%
        {wu2021fashion}
\bibfield{author}{\bibinfo{person}{Hui Wu}, \bibinfo{person}{Yupeng Gao},
  \bibinfo{person}{Xiaoxiao Guo}, \bibinfo{person}{Ziad Al-Halah},
  \bibinfo{person}{Steven Rennie}, \bibinfo{person}{Kristen Grauman}, {and}
  \bibinfo{person}{Rogerio Feris}.} \bibinfo{year}{2021}\natexlab{}.
\newblock \showarticletitle{Fashion iq: A new dataset towards retrieving images
  by natural language feedback}. In \bibinfo{booktitle}{\emph{Proceedings of
  the IEEE/CVF Conference on computer vision and pattern recognition}}.
  \bibinfo{pages}{11307--11317}.
\newblock


\bibitem[\protect\citeauthoryear{Yu, Lee, Choi, and Kim}{Yu
  et~al\mbox{.}}{2020}]%
        {yu2020curlingnet}
\bibfield{author}{\bibinfo{person}{Youngjae Yu}, \bibinfo{person}{Seunghwan
  Lee}, \bibinfo{person}{Yuncheol Choi}, {and} \bibinfo{person}{Gunhee Kim}.}
  \bibinfo{year}{2020}\natexlab{}.
\newblock \showarticletitle{Curlingnet: Compositional learning between images
  and text for fashion iq data}.
\newblock \bibinfo{journal}{\emph{arXiv preprint arXiv:2003.12299}}
  (\bibinfo{year}{2020}).
\newblock


\bibitem[\protect\citeauthoryear{Yuan and Lam}{Yuan and Lam}{2021}]%
        {yuan-2021-conversational}
\bibfield{author}{\bibinfo{person}{Yifei Yuan} {and} \bibinfo{person}{Wai
  Lam}.} \bibinfo{year}{2021}\natexlab{}.
\newblock \showarticletitle{Conversational Fashion Image Retrieval via
  Multiturn Natural Language Feedback}. In \bibinfo{booktitle}{\emph{Proc. of
  International ACM SIGIR Conference on Research and Development in Information
  Retrieval (SIGIR)}}. \bibinfo{publisher}{ACM}.
\newblock
\urldef\tempurl%
\url{https://doi.org/10.1145/3404835.3462881}
\showDOI{\tempurl}


\bibitem[\protect\citeauthoryear{Zhan, Wu, Dong, Wei, Lu, Zhang, Xu, and
  Liang}{Zhan et~al\mbox{.}}{2021}]%
        {Zhan_2021_ICCV}
\bibfield{author}{\bibinfo{person}{Xunlin Zhan}, \bibinfo{person}{Yangxin Wu},
  \bibinfo{person}{Xiao Dong}, \bibinfo{person}{Yunchao Wei},
  \bibinfo{person}{Minlong Lu}, \bibinfo{person}{Yichi Zhang},
  \bibinfo{person}{Hang Xu}, {and} \bibinfo{person}{Xiaodan Liang}.}
  \bibinfo{year}{2021}\natexlab{}.
\newblock \showarticletitle{Product1M: Towards Weakly Supervised Instance-Level
  Product Retrieval via Cross-Modal Pretraining}. In
  \bibinfo{booktitle}{\emph{Proc. of IEEE/CVF International Conference on
  Computer Vision (ICCV)}}. \bibinfo{pages}{11782--11791}.
\newblock


\bibitem[\protect\citeauthoryear{Zhao, Feng, Wu, and Yan}{Zhao
  et~al\mbox{.}}{2017}]%
        {8100135}
\bibfield{author}{\bibinfo{person}{Bo Zhao}, \bibinfo{person}{Jiashi Feng},
  \bibinfo{person}{Xiao Wu}, {and} \bibinfo{person}{Shuicheng Yan}.}
  \bibinfo{year}{2017}\natexlab{}.
\newblock \showarticletitle{Memory-Augmented Attribute Manipulation Networks
  for Interactive Fashion Search}. In \bibinfo{booktitle}{\emph{Proc. of
  Conference on Computer Vision and Pattern Recognition (CVPR)}}.
  \bibinfo{pages}{6156--6164}.
\newblock
\urldef\tempurl%
\url{https://doi.org/10.1109/CVPR.2017.652}
\showDOI{\tempurl}


\bibitem[\protect\citeauthoryear{Zheng, Yang, and Tian}{Zheng
  et~al\mbox{.}}{2017}]%
        {zheng2017sift}
\bibfield{author}{\bibinfo{person}{Liang Zheng}, \bibinfo{person}{Yi Yang},
  {and} \bibinfo{person}{Qi Tian}.} \bibinfo{year}{2017}\natexlab{}.
\newblock \showarticletitle{{SIFT} meets {CNN}: A decade survey of instance
  retrieval}.
\newblock \bibinfo{journal}{\emph{IEEE Transactions on Pattern Analysis and
  Machine Intelligence (TPAMI)}} \bibinfo{volume}{40}, \bibinfo{number}{5}
  (\bibinfo{year}{2017}), \bibinfo{pages}{1224--1244}.
\newblock


\end{thebibliography}

\end{document}